\documentclass[draft]{agujournal2019}
\usepackage{url}
\usepackage{lineno}
\usepackage{amsmath}
\usepackage{array}
\usepackage{fix-cm}
\usepackage{soul}
\usepackage{algorithm}
\usepackage{algpseudocode}
\usepackage{amssymb}
\usepackage{etoolbox}
\patchcmd{\abstract}{\newpage}{\par\addvspace{18pt}}{}{\PackageError{main}{Could not patch abstract page break}{Check the document class definition.}}
\draftfalse

\journalname{Space Weather}

\begin{document}
\makeatletter
\let\@oddhead\@empty
\let\@evenhead\@empty
\makeatother

\title{Physics-Guided Flow Matching: A Generative AI Framework for Global Total Electron Content Forecasting}

\authors{
  Si-Tong Yang\affil{1},
  Feng Ye\affil{1},
  Xian Lu\affil{2},
  Dong Lin\affil{2}, and
  Benjamin C. Martinez\affil{3}
  }

\affiliation{1}{Department of Industrial Engineering, Clemson University, Clemson, SC, USA}
\affiliation{2}{Department of Physics and Astronomy, Clemson University, Clemson, SC, USA}
\affiliation{3}{Department of Physics, Utah State University, Logan, UT, USA}

\begin{abstract}
Accurate forecasting of global total electron content (TEC) is important for space weather monitoring and the reliable operation of satellite communication and navigation systems.
We propose a physics-guided generative artificial intelligence (AI) framework for day-ahead global TEC forecasting. The framework builds on flow matching, which learns to gradually transform samples from an initial distribution into new samples that share the statistical characteristics of the target data.
The proposed model incorporates ionospheric dynamics described by a partial differential equation (PDE) into this transformation process, guiding the changes between initial and target samples using physical knowledge.
The model further combines the physical dynamics with statistical patterns captured from data to learn this transformation, adaptively adjusting their relative contributions at each location.
The model is trained and evaluated using one year of global TEC data generated by the Thermosphere Ionosphere Electrodynamics General Circulation Model (TIEGCM).
The results demonstrate that our model improves both spatial and temporal forecasting accuracy over representative benchmarks, including statistical, deep learning, and generative AI methods at all lead times and across latitudes.
The average root mean square error (RMSE) is 39.2\% lower than that of ConvLSTM and 23.8\% lower than that of the same framework without physical guidance.
The learned contribution weights vary systematically with latitude and across geographic locations, showing how the physical contribution changes across the forecast domain.
These findings demonstrate that explicit physical guidance improves generative forecasts of global TEC and highlight the value of incorporating physical dynamics into generative AI modeling.
\end{abstract}

\section*{Plain Language Summary}
The ionosphere is a region of Earth's upper atmosphere containing electrically charged particles, whose distribution is shaped by multiple physical processes, resulting in complex spatiotemporal variations. One important measure of its state is total electron content (TEC), which describes how many electrons are present along a path through the ionosphere.
In this study, we propose a generative AI method called flow matching to forecast global TEC maps up to one day in advance. The method learns from past data how to transform an initial state into a possible future state.
Physical equations guide how the model learns changes between consecutive TEC maps.
The model also combines a physical estimate of these changes with an estimate learned from data, adjusting their relative contributions by location.
We tested the model using a year of global ionospheric data and compared it with representative statistical and deep learning methods. Our model produced the lowest forecast errors at all evaluated times and across all regions. Adding physical guidance also reduced errors compared to the same framework without this guidance.
This work demonstrates the value of incorporating physical knowledge into generative AI models, providing a promising path toward more reliable ionospheric forecasting for space weather applications.

\section{Introduction}
\label{sec:introduction}

The ionosphere is a partially ionized region of Earth's upper atmosphere and an important propagation medium for transionospheric radio signals \cite{lanzerotti2001space}. Variations in its electron density alter radio-wave propagation and can degrade satellite communications and Global Navigation Satellite Systems (GNSS) performance \cite{beniguel2002global,rovira2015worldwide,ratnam2018ionospheric}.
Total electron content (TEC)
is one of the most widely used quantities for characterizing ionospheric variability and assessing its effects on radio propagation \cite{hernandez-pajares2009igs}.
The electron density distribution is governed by solar radiation, geomagnetic activity, and coupling with the lower atmosphere across multiple spatial and temporal scales \cite{liu2016variability, lu2026relative}.
During disturbed space weather conditions (e.g., solar flares and geomagnetic storms), ionospheric electron density can undergo rapid and large changes \cite{gonzalez1994geomagnetic,fullerrowell1994response,lanzerotti2001space}.
Meanwhile, lower satellite manufacturing and launch costs have enabled the deployment of large constellations, making reliable monitoring and forecasting of ionospheric evolution increasingly important for space weather applications \cite{lagunas2024lowearth}.

\begin{figure}[htbp]
    \centering
    \includegraphics[width=\linewidth]{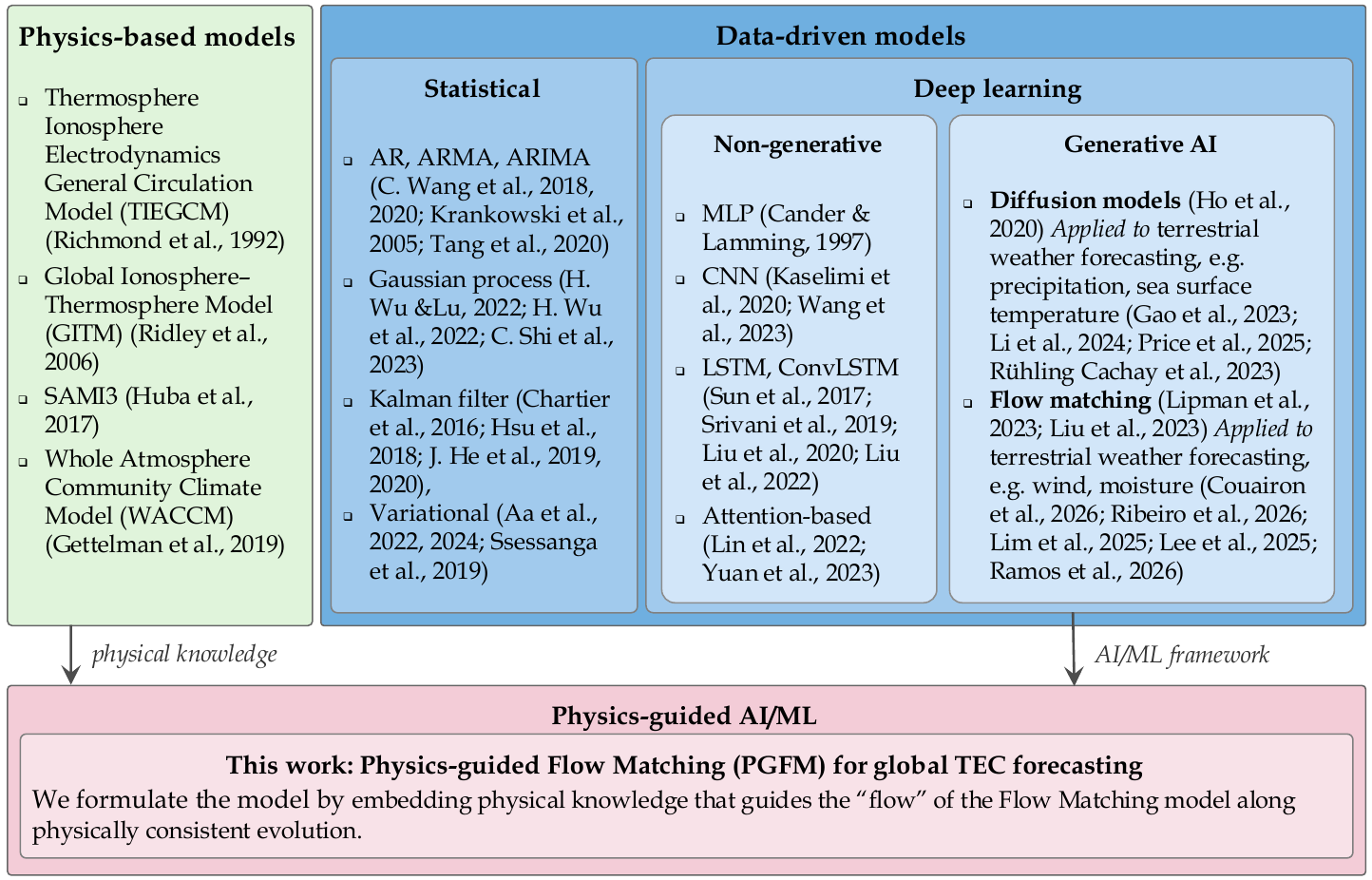}
    \caption{Overview of ionospheric modeling frameworks, organized by modeling paradigm.}
    \label{fig:methods}
\end{figure}

Approaches to ionospheric spatiotemporal modeling can generally be grouped into two main paradigms: physical models and data-driven models as summarized in Figure~\ref{fig:methods}. \textit{Physical models}, including the Thermosphere Ionosphere Electrodynamics General Circulation Model (TIEGCM) \cite{richmond1992thermosphere}, SAMI3 \cite{huba2017sami3}, the Whole Atmosphere Community Climate Model (WACCM) \cite{gettelman2019whole}, and the Global Ionosphere–Thermosphere Model (GITM) \cite{ridley2006global},
simulate ionospheric evolution by numerically solving the time-dependent continuity, momentum, and energy equations governing ionospheric dynamics.
These models provide physically interpretable representations of the system dynamics. For example, TIEGCM simulations driven by data-assimilated auroral precipitation and electric fields have been used to investigate storm-time ionosphere-thermosphere responses \cite{lu2023understanding,poudel2026ionospherethermosphere}.
TIEGCM coupled with magnetospheric models can more accurately represent the ionospheric response during geomagnetic storms by providing physically based and dynamically varying high-latitude forcing \cite{lin2022thermospheric,pham2022thermospheric}.
However, these coupled physics-based models typically require substantial computational resources \cite{hsu2023effects,mehta2018quasiphysical} and real-time solar-wind observations, which may be impractical for operational forecasting.

\textit{Data-driven methods}, including statistical and deep learning approaches, have become increasingly popular for TEC prediction.
Early statistical approaches relied largely on time series models, including autoregressive (AR) \cite{wang2018prediction, wang2020global}, autoregressive moving average (ARMA) \cite{krankowski2005wavelet}, and autoregressive integrated moving average (ARIMA) \cite{tang2020comparison} models.
In addition, Gaussian process methods have been applied to global TEC prediction and related high-latitude data assimilation \cite{wu2022data, wu2022multiresolution, shi2023predicting}. Variational techniques \cite{aa20223d, aa20243d, ssessanga2019imaging} and Kalman filter methods \cite{chartier2016ionospheric, hsu2018assessment, he2019enkf, he2020evaluation} have also supported ionospheric state estimation.
As TEC forecasting increasingly involves high-dimensional spatiotemporal data, the emergence of deep learning methods has introduced a powerful alternative by learning spatiotemporal dependencies directly from data.
Early neural network approaches were introduced for ionospheric modeling and short-term forecasting \cite{cander1997neural,hernandez1997neural}, followed by convolutional neural networks (CNNs) for TEC time-series forecasting \cite{kaselimi2020sequencetosequence} and global TEC map prediction \cite{wang2023forecast}.
Recurrent architectures, particularly LSTM networks, became widely used for modeling nonlinear temporal dependencies and forecasting TEC \cite{sun2017forecasting,srivani2019deep,liu2020forecasting}.
These architectures were extended to global TEC-map prediction \cite{liu2022ml}, multistep forecasting with encoder-decoder convolutional LSTM models \cite{xia2022edconvlstm}, and attention-enhanced spatiotemporal modeling \cite{luo2023prediction}.
More recently, Transformer-based models have further addressed spatial encoding \cite{lin2022optimal}, long-horizon forecasting \cite{yuan2023synthesisstyle}, multistep prediction \cite{shih2024forecasting}, local temporal structure \cite{li2025ionformer}, and graph-based spatiotemporal dependence \cite{yu2024graphenabled}.
These advances have improved the representation of nonlinear temporal and spatial relationships in TEC forecasting.

Generative AI models, which aim to learn the underlying distribution of observed data and generate new samples with similar statistical characteristics, have recently attracted increasing attention in scientific machine learning for spatiotemporal forecasting \cite{gao2024generative}.
Among modern generative approaches, diffusion models have emerged as a powerful tool for modeling complex high-dimensional distributions. Building on their success in generative tasks, Denoising Diffusion Probabilistic Models (DDPMs) \cite{ho2020denoisinga} have increasingly been applied to several physical forecasting tasks, including precipitation nowcasting \cite{gao2023prediff}, ensemble global weather forecasting \cite{li2024seeds}, medium-range weather forecasting \cite{price2025gencast}, and sea surface temperature forecasting \cite{ruhlingcachay2023dyffusion}.
Closely related to diffusion-based generative modeling, flow matching offers a continuous-time generative formulation that learns a ``velocity field" that defines a continuous transformation between source and target data distributions known as ``flow field" \cite{lipman2023flow}.
With a suitable ``flow", flow matching can yield straighter transform trajectories with fewer computational steps \cite{lipman2023flow,liu2023flow}.
Recent studies have explored flow matching for physical systems, including probabilistic global weather forecasting \cite{couairon2026archesweathergen}, precipitation nowcasting \cite{ribeiro2026flowcast}, spatiotemporal dynamical-system forecasting and probability-path design \cite{lim2025elucidating}, time-series forecasting of high-dimensional dynamics governed by partial differential equations (PDEs) \cite{lee2025operator}, and fluid-dynamics surrogate modeling \cite{ramos2026fluidflow}.
In summary, these studies demonstrate the growing potential of generative approaches for modeling complex spatiotemporal and physical dynamics.
Global TEC maps evolve continuously under ionospheric processes, making flow matching, which explicitly models continuous transformations, a promising framework for forecasting their spatiotemporal evolution.
However, its application to global TEC forecasting remains insufficiently explored.

Although AI-based methods are capable of capturing underlying statistical relationships in the data, they may not obey the physical principles of the systems, especially when applied to scenarios not represented in the training data \cite{kashinath2021physicsinformed}. Thus, introducing a generative formulation alone cannot guarantee that the predictions are physically consistent and scientifically sound.
This has driven an active area of research in AI and machine learning (ML) referred to as \textit{physics-informed}, or \textit{physics-guided} learning. Physics-guided learning is broadly defined as the integration of observational knowledge, physical principles, or governing laws into AI/ML models in order to guide them to adhere to certain physical aspects underpinning the driving physical process \cite{karniadakis2021physicsinformed}.
It has been studied in fluid mechanics \cite{jin2021nsfnets}, additive manufacturing \cite{zhu2021machine}, wind and energy \cite{yeAIRUWRFPhysicsGuidedSpatioTemporal2023,ye2025improved}.
For flow matching, the choice of ``flow'' shapes how the model learns to transform and can affect forecast accuracy \cite{lim2025elucidating}.
This design flexibility provides a natural opportunity to incorporate physical knowledge into the transformation as ``physical flow".
In particular, governing PDEs can guide this “physical flow” so that the learned transformations better reflect the system’s dynamics.

In this study, we propose physics-guided flow matching (PGFM), a generative AI framework for 24-hour global TEC forecasting. PGFM constructs the ``physical flow" directly following PDE-generated intermediate states. PGFM uses physical knowledge to shape the ``velocity field" target learned during training. The main contributions of this work are as follows:
\begin{itemize}
    \item Existing flow matching formulations for spatiotemporal forecasting typically construct ``flow" using generic interpolation or statistically motivated transform constructions. In contrast, we construct a physically aligned ``flow" guided by a PDE representing the intermediate evolution, thereby linking the ``flow field" to the underlying evolution of the physical system. The model then learns the ``velocity field" associated with this ``physical flow", allowing the physical guidance encoded in the path to shape the learned transport dynamics while preserving the flexibility of generative forecasting.

    \item Beyond the ``physical flow" design,
    we further incorporate ionospheric dynamics into the learning of the ``flow''.
    The model estimates these physical dynamics from historical physical information and combines them with a data-driven ``velocity field'' through a spatial weighting field learned jointly with the``velocity''estimators.
    This design enables the model to capture data-driven state transitions and the physical effects of ionospheric dynamics, with their relative contributions varying across locations.
\end{itemize}

The remainder of this paper is organized as follows. Section~\ref{sec:data_prob} describes our dataset and formulates the global TEC forecasting problem. Section~\ref{sec:method} presents the proposed methodology and model architecture. Section~\ref{sec:experiments} details the experimental setup and evaluation metrics. Section~\ref{sec:results_discussion} reports the overall and spatial forecasting results and analyzes the learned physical guidance. Finally, Section~\ref{sec:conclusion} concludes the paper and outlines directions for future work.

\section{Data Description and Problem Formulation}
\label{sec:data_prob}

\subsection{Data Description}
\label{subsec:data}

In this study, we use global ionospheric simulations generated by TIEGCM version 3.0 \cite{wu2025ncartiegcm}.
TIEGCM is a global 3D numerical model that simulates the coupled thermosphere-ionosphere system from \(\sim 97 \text{ to} \sim600~\mathrm{km}\) altitude. It self-consistently solves the fully coupled nonlinear, hydrodynamic, thermodynamic and continuity equations of the neutral gas, the ion and electron energy equations, the $\mathrm{O}^+$ continuity equation and ion chemistry, and the neutral wind dynamo \cite{roble1988coupled, richmond1992thermosphere, qian2014ncara}.
In order to incorporate lower-atmosphere variability driven by tides, planetary waves, and large-scale gravity waves, we nudge hourly SD-WACCM-X data including zonal and meridional winds, neutral temperatures, and geopotential heights at the lower levels of TIEGCM \cite{wu2023simulation, lu2025evaluatinga, lu2024impacts, lu2026relative}.

\begin{figure}[htbp]
    \centering
    \includegraphics[width=\linewidth]{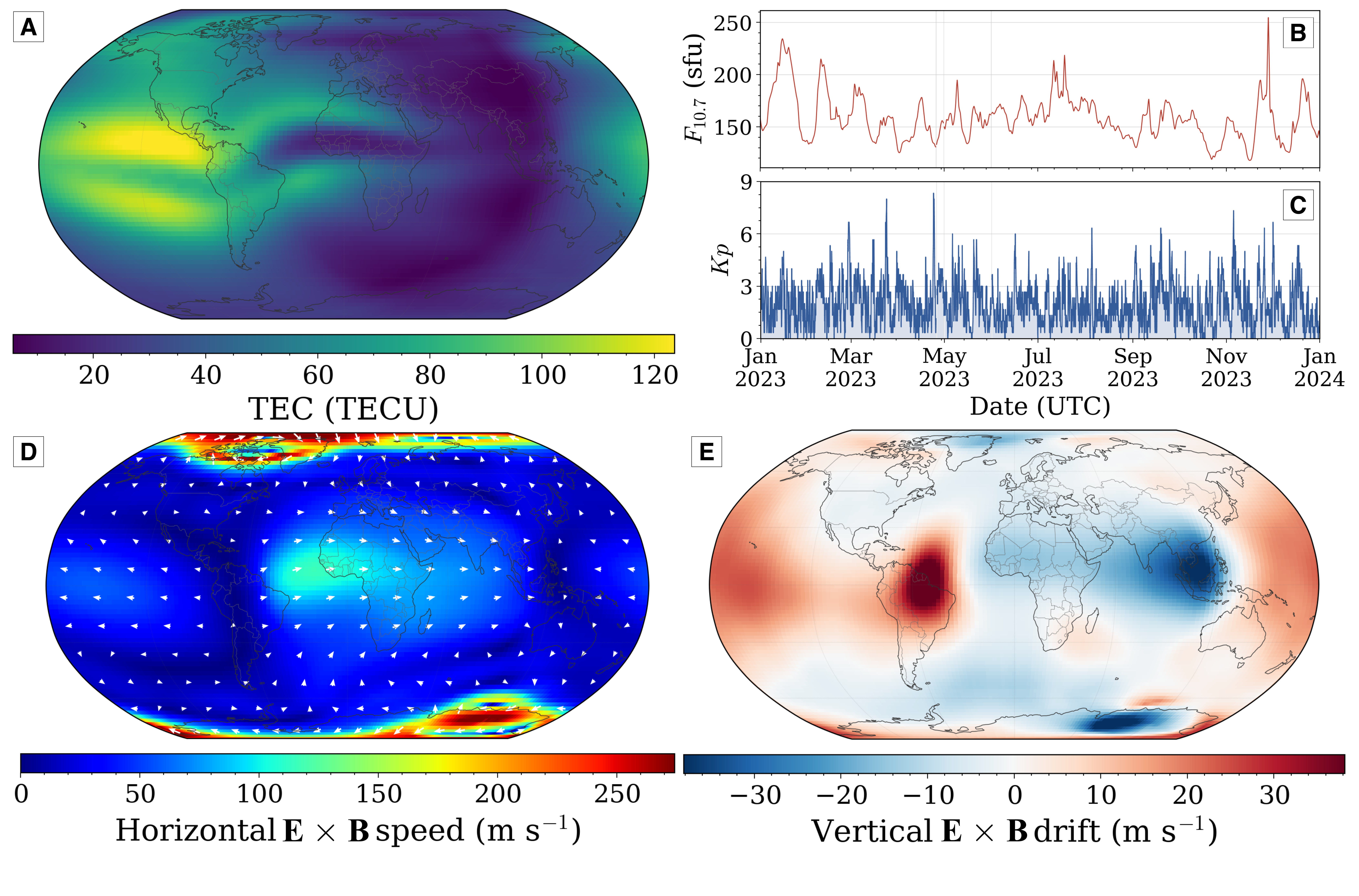}
    \caption{Overview of the dataset used in this study. (A) Global TEC field at a representative time. (B) Solar radio flux $F_{10.7}$ and (C) geomagnetic index $Kp$ over the study period. (D) Horizontal $\mathbf{E}\times\mathbf{B}$ drift field and (E) vertical $\mathbf{E}\times\mathbf{B}$ drift at the same representative time step as in (A). Colors in (D) indicate the magnitude of the horizontal drift, with arrows showing its direction, whereas colors in (E) represent the signed vertical drift velocity.
    Panels (A), (D), and (E) show global snapshots of the corresponding fields at 17 February 2023, 22:00 UTC.}
    \label{fig:data}
\end{figure}

Let \(\mathbf{s}\) denote horizontal location, \(z\) altitude, and \(t\) physical time. The model provides the three-dimensional electron number density field \(N_{\mathrm e}(\mathbf{s},z,t)\), expressed in \(\mathrm{m}^{-3}\). At each horizontal location and time, we define the TEC of the corresponding ionospheric column as Equation~\eqref{eq:tec_definition}:
\begin{equation}
X(\mathbf{s},t)
=
\mathrm{C}
\int_{z^{\mathrm{lower}}}^{z^{\mathrm{upper}}}
N_{\mathrm e}(\mathbf{s},z,t)\,\mathrm{d}z,
\quad
\mathrm{C}=10^{-16},
\label{eq:tec_definition}
\end{equation}
where \(z^{\mathrm{lower}}\) and \(z^{\mathrm{upper}}\) denote the lower and upper limits of the vertical integration domain, respectively. The lower integration boundary is located at an altitude of approximately \(97~\mathrm{km}\), whereas the upper boundary may vary with location and time. The factor \(\mathrm{C}\) converts the vertically integrated electron density from electrons per square meter to TEC units (TECU), where \(1~\mathrm{TECU}=10^{16}\) electrons per \(\mathrm{m}^{2}\). Thus, \(X(\mathbf{s},t)\) is a scalar TEC value associated with one ionospheric column. Collecting the column TEC values over the horizontal grid gives the global TEC map
\(
\mathbf{X}_t
=
\left[
X(\mathbf{s}_{ij},t)
\right]_
{\substack{i=1,\ldots,H\\j=1,\ldots,W}}
\in
\mathbb{R}^{H\times W},
\)
where \(\mathbf{s}_{ij}\) denotes the horizontal location of grid cell \((i,j)\), and \(H\) and \(W\) denote the numbers of grid cells in the latitudinal and longitudinal directions, respectively.
The TIEGCM outputs are stored every 15 min on a spatial \(2.5^{\circ}\times2.5^{\circ}\) (longitude \(\times\) latitude) grid. Each TEC field $\mathbf{X}_t$ is represented as an \(H\times W=72\times144\) gridded map, comprising 72 latitude points and 144 longitude points. Figure~\ref{fig:data} (A) presents a representative global TEC map. The experiments use one year of TIEGCM data.

In addition to the TEC data, the dataset includes several physical variables that represent diverse physical mechanisms. Figure~\ref{fig:data} (B) and (C) show that the external drivers consist of the daily solar radio flux $F_{10.7}$ and the planetary geomagnetic index $Kp$, which characterize solar and geomagnetic forcing, respectively.
The dataset also contains information derived from the dynamical equations describing the dominant ion in the F-region, such as $\mathrm{O}^+$, providing a physically interpretable characterization of electron-density variability.
We consider four physical processes: ion production, chemical loss, electrodynamic-drift transport, and neutral-wind transport. These processes contribute to the ion continuity equations shown in Figure~\ref{fig:architecture}.
As an illustrative example of the transport-related variables, Figure~\ref{fig:data} (D) and (E) present the horizontal and vertical components of the $\mathbf{E}\times\mathbf{B}$ ion drift, respectively, where $\mathbf{E}$ and $\mathbf{B}$ denote the electric and magnetic fields. These components characterize the electric-field-driven redistribution of ionospheric plasma.

\subsection{Forecasting Problem Formulation}
\label{subsec:problem}

\begin{figure}[htbp]
    \centering  \includegraphics[width=\linewidth]{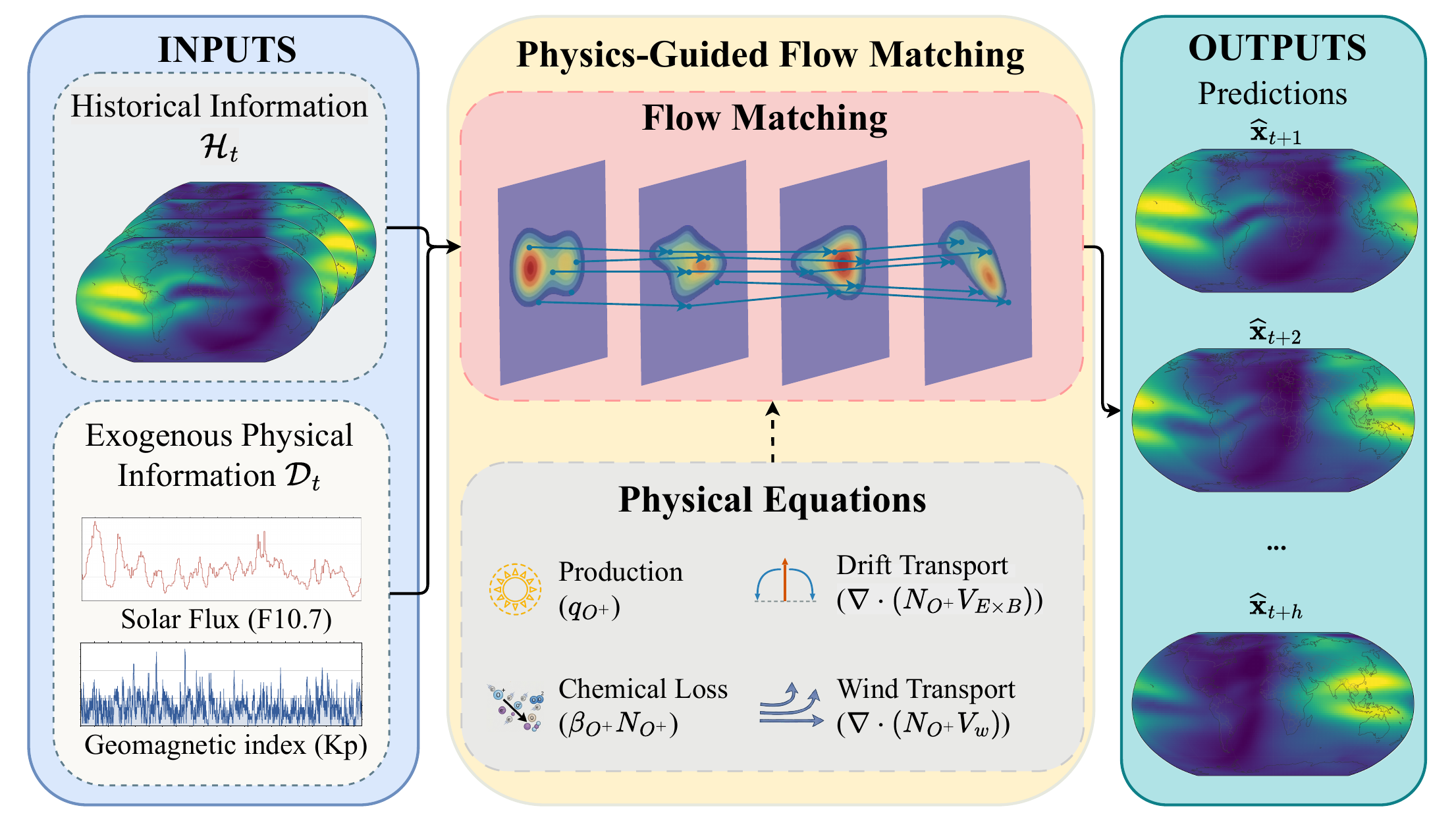}
    \caption{Overview of the proposed model for global TEC forecasting.}
    \label{fig:architecture}
\end{figure}

As illustrated in Figure~\ref{fig:architecture}, we formulate global TEC forecasting as a multistep spatiotemporal prediction problem conditioned on past TEC states, exogenous physical drivers, and ionospheric dynamics.
For a forecast initialized at time \(t\), the model uses the history
\(
\mathcal{H}_t
=
\left\{
\mathbf{X}_{t-L+1},
\ldots,
\mathbf{X}_t
\right\},
\)
where \(L\) denotes the number of historical time steps and \(\mathbf{X}_t\) is the global TEC map. The corresponding sequence of physical drivers is
\(
\mathcal{D}_t
=
\left\{
\mathbf{d}_{t-L+1},
\ldots,
\mathbf{d}_t
\right\},
\)
where \(\mathbf{d}_t\) contains the solar and geomagnetic drivers available at time \(t\). Given \(\mathcal{H}_t\), \(\mathcal{D}_t\), and the physical-process equations described in Figure~\ref{fig:architecture}, the forecasting objective is to predict the next \(h\) TEC maps, \(
{\mathbf{X}}_{t+1},
\ldots,
{\mathbf{X}}_{t+h}.
\)

\section{Methodology}
\label{sec:method}

\subsection{Flow Matching}
\label{subsec:fm}

Flow matching learns a ``velocity field" that transports samples from a prior distribution $\pi_0$ to a target data distribution $\pi_1$ \cite{lipman2023flow}. Let $\mathrm z_0\sim\pi_0$ denote an initial sample drawn from the prior distribution, and let $\mathrm z_1$ denote the generated sample at the end of the transformation, whose distribution is intended to match the target data distribution $\pi_1$.
The intermediate state is denoted by $\mathrm z_{\tau}$, where $\tau\in[0,1]$ is the normalized flow coordinate that parameterizes the flow from the prior distribution at $\tau=0$ to the target distribution at $\tau=1$.
As samples evolve from their initial states $\mathrm z_0$ toward their final states $\mathrm z_1$, their distribution also changes.
A ``probability path" $\{\pi_{\tau}\}_{\tau\in[0,1]}$ specifies the desired distribution of samples at each stage of this transformation, connecting $\pi_0$ to $\pi_1$.
To realize this transformation, flow matching learns a time-dependent ``velocity field" $v_{\theta}(\mathrm z,\tau)$, parameterized by $\theta$, that specifies the direction and rate of change for each sample in Equation~\eqref{eq:fm_flow}:
\begin{equation}
    \frac{\mathrm{d}\mathrm z_{\tau}}{\mathrm{d}\tau}
    =
    v_{\theta}(\mathrm z_{\tau},\tau),
    \quad \tau\in[0,1].
    \label{eq:fm_flow}
\end{equation}
With initial condition $\mathrm{z}_{0}\sim\pi_{0}$, integrating Equation~\eqref{eq:fm_flow} from $0$ to $\tau$ gives
\begin{equation}
    \mathrm{z}_{\tau}
    =
    \mathrm{z}_{0}
    +
    \int_{0}^{\tau}
    v_{\theta}(\mathrm{z}_{s},s)\,\mathrm{d}s
    =
    \phi_{\tau}(\mathrm{z}_{0}),
    \label{eq:fm_flow_map}
\end{equation}
where $\phi_{\tau}$ denotes the ``flow map" generated by the trained ``velocity field". Thus, $\mathrm{z}_{\tau}$ represents the state reached by evolving the initial sample $\mathrm{z}_{0}$ under the learned ``velocity field'' $v_{\theta}$ from flow coordinate $0$ to $\tau$.
Training aims to make these evolving samples follow the designed distributions $\mathrm z_{\tau}\sim\pi_{\tau}$ and match the target distribution $\pi_1$ at $\tau=1$.
\ref{app:fm_background} provides the full conditional flow matching formulation.

\subsection{Physics-Guided Flow Matching}
\label{subsec:PGFM}
\subsubsection{Physics-Guided Probability Path}
\label{subsubsec:path_design}

To train the ``velocity field", flow matching first specifies two components: a coupling between the endpoint distributions and a conditional probability path connecting each pair of endpoint samples. The coupling specifies how samples from the two endpoint distributions are paired, and the conditional probability path describes the intermediate distributions connecting each paired sample. These choices determine the target conditional ``velocity field" and give rise to different flow matching formulations.
Given an endpoint pair $(\mathrm z_0,\mathrm z_1)\sim\gamma$, where $\gamma$ denotes the coupling between the source and target distributions, the conditional probability path $p_{\tau}(\mathrm z\mid\mathrm z_0,\mathrm z_1)$ specifies the distribution of the intermediate state $\mathrm z$ at flow time $\tau$.
The basic conditional flow matching formulation adopts independent sampling of the source and target states to form the coupling, while the intermediate states are generated through a predefined conditional probability path between the paired endpoints \cite{lipman2023flow}.

\begin{figure}[htbp]
    \centering
    \includegraphics[width=\linewidth]{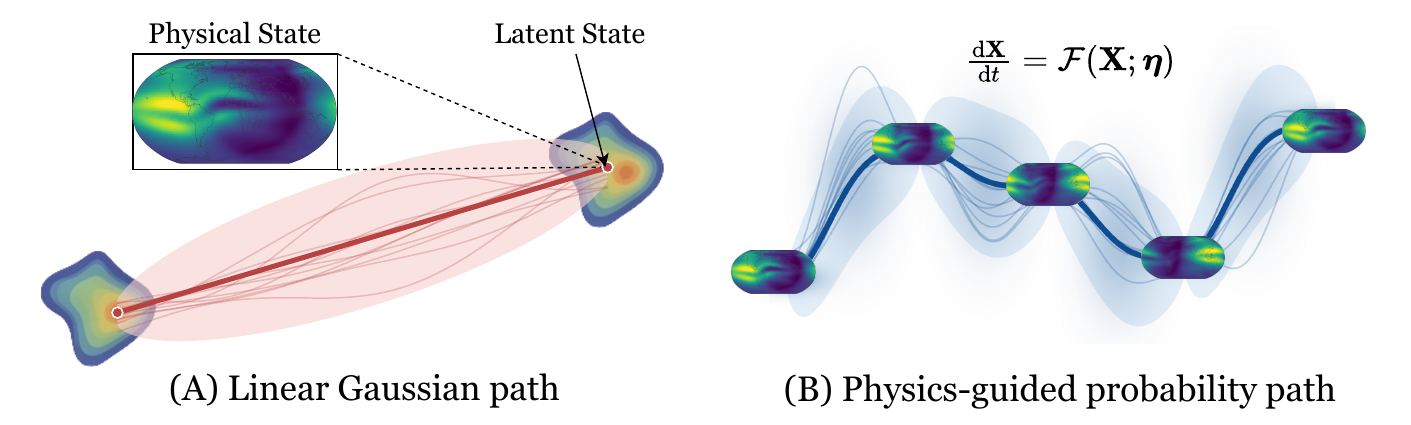}
    \caption{Illustration of the difference between (A) a linear Gaussian probability path and (B) the proposed physics-guided probability path. The solid curve denotes the conditional mean path, the thin curves represent possible stochastic trajectories, and the shaded region indicates the dispersion defined by the variance schedule.}
    \label{fig:fm_path}
\end{figure}

For spatiotemporal forecasting, temporal continuity provides a natural coupling between endpoint states. Building on this observation, the remaining design choice is the conditional probability path.
Several studies have adopted Gaussian conditional probability paths \cite{chen2024probabilistic, lim2025elucidating}, which provide a flexible modeling approach. The common form of the family of Gaussian conditional probability path $p_\tau$ could be defined as Equation~\eqref{eq:general_path}:
\begin{equation}
    p_{\tau}
    \left(
        \mathrm z \mid \mathrm z_0,\mathrm z_1
    \right)
    =
    \mathcal{N}
    \left(\mathrm z\mid\
        a_\tau \mathrm z_0+b_\tau \mathrm z_1,\,
        c^{2}_\tau\mathbf{I}
    \right).
    \label{eq:general_path}
\end{equation}
where $a_\tau$ and $b_\tau$
specify the interpolation of the conditional mean and ensure that the conditional mean equals $\mathrm z_0$ at $\tau=0$ and $\mathrm z_1$ at $\tau=1$,
while $c_\tau^2$ controls the conditional variance.
Within this family, \citeA{chen2024probabilistic} construct an asymmetric stochastic interpolant between consecutive states. \citeA{lim2025elucidating} formulate flow matching in a learned latent space and motivate a Gaussian probability path from a Schrödinger bridge perspective  \cite{leonard2013survey}. In the latter formulation, an encoder maps each observed state to a latent representation, the probability path and ``velocity field" are defined in this latent space, and a decoder maps the terminal latent state back to the observation space. As illustrated in Figure~\ref{fig:fm_path}(A),
the latter formulation constructs a conditional probability path between two latent states. The conditional mean interpolates linearly between the two endpoint states, while the conditional variance follows a symmetric schedule that is smallest at the endpoints and largest at the midpoint. This produces stochastic intermediate states around the linear mean path.

Motivated by their work, we adopt a similar coupling principle but construct the probability path directly in the observation space,
which allows our model to incorporate physical dynamics into the architecture. Specifically, rather than constructing an artificial interpolation between the endpoints, we define the physics-guided path according to the dynamical evolution driven by a PDE.
The guided flow, therefore, follows the underlying physical evolution, and we refer to it as ``physical flow".

For the ionosphere, the underlying physical processes governing the electron concentration \(N_{\mathrm e}\) can be broadly classified into two categories: photochemical processes, which determine the plasma production \(P_{\mathrm e}\) and loss \(L_{\mathrm e}\), and transport processes, represented by the electron flux
$N_{\mathrm e}\mathbf V_{\mathrm e}$, which redistribute the plasma \cite{rishbeth1969introduction,kelley2009earth}. The temporal evolution of \(N_{\mathrm e}\) is described by Equation~\eqref{eq:ne_continuity},
\begin{equation}
\frac{\partial N_{\mathrm e}}{\partial t}
=
\underbrace{P_{\mathrm e}}_{\mathrm{Production}}
-
\underbrace{L_{\mathrm e}}_{\mathrm{Loss}}
-
\underbrace{\nabla\cdot
\left(
N_{\mathrm e}\mathbf V_{\mathrm e}
\right)}_{\mathrm{Transport}}.
\label{eq:ne_continuity}
\end{equation}
Using the TEC definition in Equation~\eqref{eq:tec_definition}, we obtain the evolution equation for column TEC by vertically integrating Equation~\eqref{eq:ne_continuity}.
Horizontal and vertical plasma transport contribute differently to the column TEC balance. Decomposing the plasma velocity as $\mathbf V_{\mathrm e}=(\mathbf V_{\mathrm e,h},V_{\mathrm e,z})$,
the transport term of Equation~\eqref{eq:ne_continuity} can be expressed as Equation~\eqref{eq:transport_decomposition},
\begin{equation}
\nabla\cdot
\left(
N_{\mathrm e}\mathbf V_{\mathrm e}
\right)
=
\nabla_h\cdot
\left(
N_{\mathrm e}\mathbf V_{\mathrm e,h}
\right)
+
\frac{\partial}{\partial z}
\left(
N_{\mathrm e}V_{\mathrm e,z}
\right),
\label{eq:transport_decomposition}
\end{equation}
where $\nabla_h\cdot$ denotes the horizontal divergence operator.
For vertical integration, we define
$Z=[z^{\mathrm{lower}},z^{\mathrm{upper}}]$
and, for brevity, write $\int_Z(\cdot)\,\mathrm{d}z$ for integration over this interval.
We treat the lower boundary as fixed and neglect the correction terms that arise, by the Leibniz rule, from the temporal and horizontal variation of the upper boundary, assuming that they are small relative to the retained terms.
Under this approximation, the TEC balance becomes
\begin{equation}
\frac{\partial X}{\partial t}
\approx
\mathrm{C}\int_Z P_{\mathrm e}\,\mathrm{d}z
-
\mathrm{C}\int_Z L_{\mathrm e}\,\mathrm{d}z
-
\mathrm{C}\nabla_h\cdot
\left(
\int_Z N_{\mathrm e}\mathbf V_{\mathrm e,h}\,\mathrm{d}z
\right)
-
\mathrm{C}\left[
N_{\mathrm e}V_{\mathrm e,z}
\right]_{z^{\mathrm{lower}}}^{z^{\mathrm{upper}}}.
\label{eq:tec_integrated_exact}
\end{equation}
Collecting Equation~\eqref{eq:tec_integrated_exact} over the horizontal grid gives the evolution of the global TEC map which depends on time only. We write these map-level dynamics in the compact form of Equation~\eqref{eq:pde_general}:
\begin{equation}
\frac{\mathrm{d}\mathbf{X}}{\mathrm{d}t}
=
\mathcal{F}
\left(
\mathbf{X};
\boldsymbol{\eta}
\right),
\label{eq:pde_general}
\end{equation}
where $\mathcal{F}$ is the physical evolution operator that evaluates the right-hand side of Equation~\eqref{eq:tec_integrated_exact} at every grid column, and $\boldsymbol{\eta}$ represents the associated physical conditions, such as solar irradiance, geomagnetic activity, neutral winds, and electric fields.
The PDE in Equation~\eqref{eq:pde_general} defines the continuous evolution of the physical state. Let $\mathbf{X}_t$ and $\mathbf{X}_{t+1}$ denote two consecutive observed TEC states with a time gap $\Delta t$, corresponding to the endpoint states $\mathbf{x}_0$ and $\mathbf{x}_1$ in the flow matching formulation, respectively.
We use the physical evolution of this simulated transition to construct a physics-guided path over the flow coordinate $\tau\in[0,1]$.
The conditional mean of the PDE-guided path is then defined as shown in Equation~\eqref{eq:pde_mean}:
\begin{equation}
\boldsymbol{\mu}_{\mathrm{PDE}}(\tau)
=
\mathbf{x}_0
+
\int_{t}^{t+\tau\Delta t}
\mathcal{F}\left(\mathbf{X}_r;
\boldsymbol{\eta}\right)
\,\mathrm{d}r,
\label{eq:pde_mean}
\end{equation}
where $\Delta t$ denotes the physical time gap between the two consecutive observations, $r$ denotes physical time, and $\mathbf X_r$ denotes the physical trajectory represented by the TIEGCM simulation. Equation~\eqref{eq:pde_mean} gives the continuous dynamical representation of this trajectory. In training, we approximate the mean path by piecewise-linear interpolation of the intermediate simulated states from the same transition, so that $\boldsymbol{\mu}_{\mathrm{PDE}}(0)=\mathbf{x}_0$ and $\boldsymbol{\mu}_{\mathrm{PDE}}(1)=\mathbf{x}_1$. The flow coordinate $\tau$ corresponds to an elapsed physical time of $\tau\Delta t$.

Using this PDE-evolved state as the conditional mean, and following the symmetric noise schedule, the resulting physics-guided probability path is then defined as shown in Equation~\eqref{eq:PDE_path}:
\begin{equation}
p_{\tau}
\left(
\mathbf{x}
\mid
\mathbf{x}_0,
\mathbf{x}_1,
\mathcal{F},
\boldsymbol{\eta}
\right)
=
\mathcal{N}
\left(
\mathbf{x}
\;\middle|\;
\boldsymbol{\mu}_{\mathrm{PDE}}(\tau),
\left[
\sigma_{\min}^2
+
\sigma^2\tau(1-\tau)
\right]\mathbf{I}
\right),
\label{eq:PDE_path}
\end{equation}
where $\sigma_{\min}^2$ is a small variance floor that avoids a degenerate conditional distribution at the endpoints, and $\sigma$ controls the magnitude of the stochastic perturbation along the path. The variance floor smooths the endpoint distributions around $\mathbf{x}_0$ and $\mathbf{x}_1$ without changing the endpoint means.
The symmetric term $\sigma^2\tau(1-\tau)$ is largest at $\tau=0.5$ and decreases toward both endpoints. This variance schedule provides stochastic smoothing around the PDE-guided trajectory for flow matching training. Under this construction, the PDE determines the mean evolution of the probability path, while the designed variance controls the dispersion of the intermediate training states around that physics-guided evolution.
This physics-guided construction defines a valid probability path for flow matching \cite{lipman2023flow,tong2024improving}.

Figure~\ref{fig:fm_path} illustrates the distinction between the two constructions. As shown in Figure~\ref{fig:fm_path} (A), the conditional probability path is constructed by linearly interpolating between two endpoints in latent space, with a Gaussian distribution introduced along the path. Each point on the path corresponds to a latent representation of the physical data state.
Figure~\ref{fig:fm_path} (B) schematically illustrates the proposed construction, in which the probability path is defined directly in the TEC-map state space. Its conditional mean $\boldsymbol{\mu}_{\mathrm{PDE}}$ evolves from the initial TEC field according to the PDE operator $\mathcal{F}$, so that the intermediate states remain interpretation as spatial TEC fields. Physically, this evolution can represent organized changes in the TEC distribution arising from photochemical production and chemical loss as well as plasma transport. The proposed construction is intended to replace the linear interpolation widely used in the literature with a probability path guided by the modeled physical evolution.

\subsubsection{Adaptive ``Velocity" Fusion with Physical Dynamics}
\label{subsubsec:physics_conditioning}

Flow matching learns a ``velocity" that transports samples from the source distribution to the target distribution along the probability path. The target ``velocity" is generated by sampling the random intermediate states $\mathbf{x}_{\tau}$ on the predefined probability path. The derivative of the sampled state with respect to $\tau$ provides the conditional target ``velocity" associated with that training state. By regressing these conditional target velocities over different endpoint pairs, flow coordinates, and stochastic perturbations, the model learns the ``velocity" corresponding to the designed probability path.
During forecasting, integrating the learned ``velocity" from $\tau=0$ to $\tau=1$ progressively transports an initial sample toward the target distribution.

For the physics-guided probability path in Equation~\eqref{eq:PDE_path}, an intermediate state is sampled using
\begin{equation}
\mathbf{x}_{\tau}
=
\boldsymbol{\mu}_{\mathrm{PDE}}(\tau)
+
s_{\tau}\boldsymbol{\epsilon},
\qquad
s_{\tau}
=
\left[
\sigma_{\min}^2+\sigma^2\tau(1-\tau)
\right]^{1/2},
\qquad
\boldsymbol{\epsilon}\sim\mathcal{N}(\mathbf{0},\mathbf{I}).
\end{equation}
For a fixed realization of $\boldsymbol{\epsilon}$, differentiating this sampled trajectory with respect to $\tau$ gives its conditional target ``velocity":
\begin{equation}
\begin{aligned}
\mathbf{u}^{\mathrm{target}}
\left(\mathbf{x}_{\tau},\tau\right)
&=
\frac{\mathrm{d}\mathbf{x}_{\tau}}{\mathrm{d}\tau}
\\
&=
\Delta t\,
\mathcal{F}\left(
\mathbf{X}_{t+\tau\Delta t};
\boldsymbol{\eta}
\right)
+
\frac{
\sigma^2(1-2\tau)
}{
2\left[
\sigma_{\min}^2+\sigma^2\tau(1-\tau)
\right]^{1/2}
}
\boldsymbol{\epsilon}.
\end{aligned}
\label{eq:physics_target_velocity}
\end{equation}
The first term describes the evolution of the PDE-guided mean, while the second arises from the variation of the stochastic dispersion along the path. The model is trained to approximate these conditional target velocities at randomly sampled intermediate states and flow coordinates.

Beyond this construction, physical information is also incorporated during training through a physical estimator, which complements the data-driven estimator. The two estimators guide the learning of the ``velocity". The data-driven estimator captures statistical dependencies in the dataset, while the physical estimator provides guidance grounded in the physical dynamics.
The physical estimator constructs a flow matching ``velocity" from the ionospheric dynamics with an explicit physical interpretation. Although this estimator is trained using a neural network, it forecasts the net TEC tendency represented by the combined $\mathrm{O}^+$ process contributions rather than directly estimating the flow matching velocity. This choice is motivated by our analysis of the TIEGCM data, which indicates that changes in $\mathrm{O}^+$ density account for most of the electron-density tendency. The $\mathrm{O}^+$ density dynamics reflect the combined effects of ion production, chemical loss, electrodynamic drift, and neutral-wind transport. These processes describe the spatiotemporal evolution of $\mathrm{O}^+$ density and provide a physical basis for modeling electron-density changes and the corresponding TEC evolution.
To represent the physical-process history used by this estimator, we define
\(
\mathcal{P}_t
=
\left\{
\mathbf{p}_{t-L+1},
\ldots,
\mathbf{p}_t
\right\},
\)
where \(\mathbf{p}_t\) collects the gridded contributions from ion production, chemical loss, electrodynamic-drift transport, and neutral-wind transport at time \(t\).
We introduce a causal physics forecaster $g_{\psi}$, parameterized by $\psi$ and implemented using a Convolutional Gated Recurrent Unit (ConvGRU) \cite{ballas2016delving}. The encoder summarizes the past ion dynamics $\mathcal{P}_{t}$ and the available physical drivers $\mathcal{D}_{t}$ into a set of hidden states. The decoder recurrently evolves the hidden states and produces a sequence of net TEC tendency estimates over one forecast cycle,
\(
\left\{
\widehat{\mathbf r}_{\psi}
\right\}
=
g_{\psi}
\left(
\mathcal{P}_{t},
\mathcal{D}_{t}
\right).
\)
The predicted net tendencies $\widehat{\mathbf r}_{\psi}$ are expressed in physical units, whereas the ``velocity" is defined over the normalized state space and flow coordinate. We then transform the predicted dynamics into the ``velocity", \(\mathbf{u}^{\mathrm{phys}}_{\psi}(\tau)
=
\mathcal{S}
\left(
\widehat{\mathbf r}_{\psi}
\right),\)
where $\mathcal{S}$ maps the predicted physical dynamics to the normalized flow ``velocity". The resulting vector \(\mathbf{u}^{\mathrm{phys}}_{\psi}\) serves as the physical ``velocity". Details of this estimator are provided in Section~\ref{app:physics_velocity_network} of \ref{app:networks}.

In parallel with the physical estimator, the data-driven estimator learns a ``velocity'' from the observed system evolution. Following the standard conditional flow matching formulation, the ``velocity" network $f_{\theta}$, parameterized by $\theta$, takes the intermediate flow state $\mathbf{x}_{\tau}$, and the normalized flow coordinate $\tau$ as its basic inputs.
To incorporate information relevant to the forecast context, we further condition the network on the historical state sequence $\mathcal{H}_{t}$ and the available physical drivers $\mathcal{D}_{t}$, yielding \(\mathbf{u}^{\mathrm{data}}_{\theta}
=
f_{\theta}
\left(
\mathbf{x}_{\tau},
\tau,
\mathcal{H}_{t},
\mathcal{D}_{t}
\right).\)
The historical TEC fields $\mathcal{H}_{t}$ and physical drivers $\mathcal{D}_{t}$ are convolutionally encoded and then concatenated with state $\mathbf{x}_{\tau}$ and the flow coordinates $\tau$. The combined input tensor is then mapped by a conditional U-Net \cite{ronneberger2015unet} to the data-driven ``velocity" $\mathbf{u}_{\theta}^{\mathrm{data}}$.

The final ``velocity" adaptively combines the data-driven and physical ``velocity'' estimates through a learned weighting map $\boldsymbol{\lambda}(\tau)\in[0,1]^{H\times W}$, as shown in Equation~\eqref{eq:fused_velocity}:
\begin{equation}
\widehat{\mathbf{u}}_{\theta,\psi,\lambda}
=
\left(
\mathbf{1}
-
\boldsymbol{\lambda}
\right)
\odot
\mathbf{u}^{\mathrm{data}}_{\theta}
+
\boldsymbol{\lambda}
\odot
\mathbf{u}^{\mathrm{phys}}_{\psi},
\label{eq:fused_velocity}
\end{equation}
The weighting map depends on the current flow state and forecast context, allowing the relative contributions of the two estimates to vary across spatial locations and along the flow coordinate $\tau$.
During training, the objective of flow matching is applied to the fused velocity. The fusion weights are learned to reduce the discrepancy between this final estimate and the target velocity. The network architectures are detailed in \ref{app:networks}; the alignment and fusion of the two``velocity''estimates are described in Section~\ref{app:adaptive_velocity_fusion}.
At each spatial location, values of $\boldsymbol{\lambda}$ close to one assign greater weight to the physical ``velocity", whereas values close to zero favor the data-driven ``velocity".
Algorithm~\ref{alg:pgfm_training} summarizes the joint training procedure.

\begin{algorithm}[t]
\caption{Training Procedure for Physics-Guided Flow Matching}
\label{alg:pgfm_training}
\begin{algorithmic}[1]
\Require Historical TEC window $\mathcal{H}_t$, exogenous physical covariates $\mathcal{D}_t$, historical ion-dynamic terms $\mathcal{P}_t$, probability path $p_\tau$
\State Sample $\tau \sim \mathcal{U}(0,1)$
\State Construct $\mathbf{x}_{\tau}$ and $\mathbf{u}^{\mathrm{target}}$ from the probability path $p_\tau$
\State Learn the ``velocity":
$\widehat{\mathbf{u}}_{\theta,\psi,\lambda}
=
\left(
1-\mathbf{\lambda}
\right)
\odot
\mathbf{u}^{\mathrm{data}}_{\theta}
+
\mathbf{\lambda}
\odot
\mathbf{u}^{\mathrm{phys}}_{\psi}$,
\State Optimize $
    \min_{\theta,\psi,\lambda}
    \left\|
    \widehat{\mathbf{u}}_{\theta,\psi,\lambda}
    -
    \mathbf{u}^{\mathrm{target}}
    \right\|^2$
\end{algorithmic}
\end{algorithm}

The fused ``velocity" is trained to match the target ``velocity" defined by the physics-guided probability path. For each training transition, we sample $\tau\sim\mathcal{U}(0,1)$ and $\boldsymbol{\epsilon}\sim\mathcal{N}(\mathbf{0},\mathbf{I})$, and construct $\mathbf{x}_{\tau}$ and $\mathbf{u}^{\mathrm{target}}(\mathbf{x}_{\tau},\tau)$ using Equation~\eqref{eq:physics_target_velocity}. Following the flow matching objective of \citeA{lipman2023flow}, we define the training loss as Equation~\eqref{eq:pgfm_training_objective}:
\begin{equation}
\mathcal{L}_{\mathrm{FM}}(\theta,\psi,\lambda)
=
\mathbb{E}_{t,\tau,\boldsymbol{\epsilon}}
\Big[
\big\|
\widehat{\mathbf{u}}_{\theta,\psi,\lambda}
\left(
\mathbf{x}_{\tau},
\tau,
\mathcal{H}_{t},
\mathcal{D}_{t},
\mathcal{P}_{t}
\right)
-
\mathbf{u}^{\mathrm{target}}
\left(
\mathbf{x}_{\tau},
\tau
\right)
\big\|_2^2
\Big].
\label{eq:pgfm_training_objective}
\end{equation}
The parameters $\theta$ and $\psi$ of the data-driven and physical estimators, with the adaptive fusion parameter $\boldsymbol{\lambda}$, are optimized jointly through Equation~\eqref{eq:pgfm_training_objective}. By matching the predicted and target velocities at randomly sampled flow coordinates $\tau$, the model learns the ``velocity" associated with the physics-guided probability path.

\subsubsection{Forecast}
\label{sec:forecast}

At forecast time $t$, given the current TEC field $\mathbf{X}_{t}$, the historical TEC window $\mathcal{H}_{t}$, the history of physical drivers $\mathcal{D}_{t}$, and the ion-dynamic history $\mathcal{P}_{t}$, the one-step forecast is obtained by integrating the learned ``velocity" over the flow coordinate $\tau\in[0,1]$.
And we initialize the flow state from the current TEC field, $\mathbf{x}_{0}={\mathbf{X}_{t}}$,
and it evolves according to Equation~\eqref{eq:forecast_ode},
\begin{equation}
    \frac{\mathrm{d}\mathbf{x}_{\tau}}{\mathrm{d}\tau}
    =
    \widehat{\mathbf{u}}_{\theta,\psi,\lambda}
    \left(
        \mathbf{x}_{\tau},
        \tau,
        \mathcal{H}_{t},
        \mathcal{D}_{t},
        \mathcal{P}_{t}
    \right).
    \label{eq:forecast_ode}
\end{equation}
The terminal flow state $\mathbf{x}_1$ represents the one-step forecast:
\begin{equation}
    \mathbf{x}_1
    =
    \mathbf{x}_0
    +
    \int_0^1
    \widehat{\mathbf{u}}_{\theta,\psi,\lambda}
    \left(
        \mathbf{x}_\tau,
        \tau,
        \mathcal{H}_t,
        \mathcal{D}_t,
        \mathcal{P}_t
    \right)
    \, d\tau .
\end{equation}
The corresponding forecast in the original TEC space is obtained by
\(
    \widehat{\mathbf{X}}_{t+1}
    =
    \mathbf{x}_1.
\)

For multi-step forecasting, the same one-step procedure is applied autoregressively. After obtaining $\widehat{\mathbf{X}}_{t+1}$, the predicted TEC field is appended to $\mathcal{H}_{t}$ and the oldest TEC field is removed. The resulting historical window is then used to predict $\widehat{\mathbf{X}}_{t+2}$. At each forecast cycle, the flow coordinate is reset to $\tau=0$. Repeating this procedure produces the
forecast sequence \(\left\{
        \widehat{\mathbf{X}}_{t+1},
        \ldots,
        \widehat{\mathbf{X}}_{t+h}
    \right\}.\)

\section{Experiments}
\label{sec:experiments}

\subsection{Experiment Design}
\label{subsec:experiment_design}

We aim to forecast hourly TEC maps with a maximum forecasting window of 1 day. Each TEC field represents a state, and consecutive hourly states define the endpoints of a flow transition. We leverage the previous 24 hourly TEC maps with the corresponding historical physical information available at forecast initialization.
The model is trained to learn a 1-hour conditional transition and is applied autoregressively for multi-hour forecasting. It learns a conditional ``velocity" over $\tau\in[0,1]$ by matching the target``velocity''defined by the physics-guided probability path. During inference, the learned ``velocity" is integrated to produce the TEC forecast 1~hour ahead. The prediction is then appended to the history window, the oldest state is removed, and the updated window is used for the subsequent prediction. This process is repeated until the 24-hour forecast horizon is reached.

The model was developed and evaluated using one year (365 days) of global TEC data generated by TIEGCM. For each month, three days were held out for testing and two days for validation, while all remaining days were used for training. This resulted in 36 test days, 24 validation days, and 305 training days.
All experiments were conducted
using two NVIDIA A100 GPUs, each with 80~GB of memory. Model parameters are optimized using the flow matching objective defined in Equation~\eqref{eq:pgfm_training_objective}. Each training run required approximately 6~hours, while generating a one-day forecast required only about 20 seconds.

We compare the proposed Physics-Guided Flow Matching (PGFM) model with representative benchmarks spanning vanilla flow matching, deep learning, and statistical methods. These models are included to assess PGFM against closely related established data-driven forecasting approaches. In addition, we include an oracle variant with access to future physical information to assess the potential performance gain provided by the physical-prior branch:
\begin{itemize}
    \item \textbf{Spatiotemporal Gaussian Process (STGP).}
    STGP is a statistical benchmark that models the spatial and temporal dependence of TEC through a Gaussian process. It produces forecasts based on the same prediction setting used for model comparison. We implement STGP using GPyTorch \cite{gardner2018gpytorch} as a sparse variational Gaussian process following the inducing-point formulation of \citeA{hensman2013gaussian}.

    \item \textbf{Convolutional Long Short-Term Memory (ConvLSTM).}
    ConvLSTM is a recurrent deep learning benchmark for gridded spatiotemporal TEC forecasting. It receives the same historical TEC fields and exogenous variables as PGFM and generates hourly forecasts autoregressively over the 24-hour forecast horizon. We implement the ConvLSTM benchmark in PyTorch following the convolutional recurrent formulation introduced by \citeA{shi2015convolutional}.

    \item \textbf{Flow Matching Baseline (FMB).}
    This baseline uses the same flow matching framework, historical conditioning inputs, and autoregressive forecast protocol used by PGFM but removes the physics-guided components. The comparison between this baseline and PGFM measures the combined contribution of the proposed physics-guided components within the same generative forecasting framework.

    \item \textbf{Operational PGFM.}
    This is our proposed framework. During forecasting, it uses only information available at the forecast initialization and does not have access to the true future physical quantities.

    \item \textbf{PGFM with True Future Physical Dynamics (Oracle).}
    This diagnostic configuration uses the same architecture as Operational PGFM but is supplied with the true future physical dynamics during forecasting. Such information is unavailable in real-time operational settings; therefore, Oracle PGFM is not deployable and is excluded from model ranking. Comparing the two configurations isolates the effect of access to accurate future physical information and estimates the potential performance gain that could be achieved if these physical quantities were predicted more accurately.
\end{itemize}
All deployable models use only information available at the forecast initialization time, ensuring a realistic operational forecasting setting. Oracle, which has access to future physical information, is excluded from the operational performance comparison and is only used for diagnostic purposes.

\subsection{Evaluation Metrics}
\label{subsec:evaluation_metrics}

For a forecast initialized at hour $t$, the prediction at lead time $h$ is valid at hour $t+h$. We use the corresponding TEC map generated by TIEGCM at this valid time as the reference. Let $\mathcal{G}$ denote the set of valid spatial grid cells, with $N=|\mathcal{G}|$. For each grid cell $i\in\mathcal{G}$, the predicted TEC value is denoted by $\widehat{\mathrm{X}}_{t+h,i}$, while the corresponding TIEGCM reference value is denoted by $\mathrm{X}_{t+h,i}$. We evaluate each forecast map using root mean square error (RMSE), mean absolute error (MAE), and normalized mean absolute error (nMAE):
\begin{equation}
\begin{aligned}
\mathrm{RMSE}_{t,h}
&=
\sqrt{
\frac{1}{N}
\sum_{i\in\mathcal{G}}
\left(
\widehat{\mathrm X}_{t+h,i}-\mathrm X_{t+h,i}
\right)^2
},
\\
\mathrm{MAE}_{t,h}
&=
\frac{1}{N}
\sum_{i\in\mathcal{G}}
\left|
\widehat{\mathrm X}_{t+h,i}-\mathrm X_{t+h,i}
\right|,
\\
\mathrm{nMAE}_{t,h}
&=
100\times
\frac{
\sum_{i\in\mathcal{G}}
\left|
\widehat{\mathrm X}_{t+h,i}-\mathrm X_{t+h,i}
\right|
}{
\sum_{i\in\mathcal{G}}
\left|
\mathrm X_{t+h,i}
\right|
}.
\end{aligned}
\label{eq:metrics}
\end{equation}

\section{Results and Discussion}
\label{sec:results_discussion}

This section evaluates the forecasting performance of the proposed PGFM model and examines the physical dynamics represented by its learned ``velocity". Here, the lead time denotes the elapsed time between forecast initialization and the valid time of a predicted TEC map, whereas the forecast horizon denotes the maximum lead time considered, which is 24~hours in this study. We first compare the global TEC maps predicted by PGFM with the corresponding TIEGCM reference maps at selected lead times to assess how well PGFM reproduces the spatial evolution of TEC. We then examine continuous latitude-resolved profiles of RMSE, MAE, and nMAE to characterize the spatial structure of the forecast errors and relate these errors to the background TEC distribution. Next, we quantify the overall forecasting performance of PGFM over the 24-hour forecast horizon by comparing it with STGP, ConvLSTM, and FMB. We also include Oracle as a diagnostic reference to assess the potential benefit of having access to true future physical information. Finally, we analyze the learned ``velocity" to investigate how the physics guidance shapes the modeled TEC evolution.

\subsection{Global TEC Forecasts and Spatial Error Patterns}
\label{subsec:spatial_error_distribution}

\begin{figure}[t]
\centering
\includegraphics[width=\textwidth]{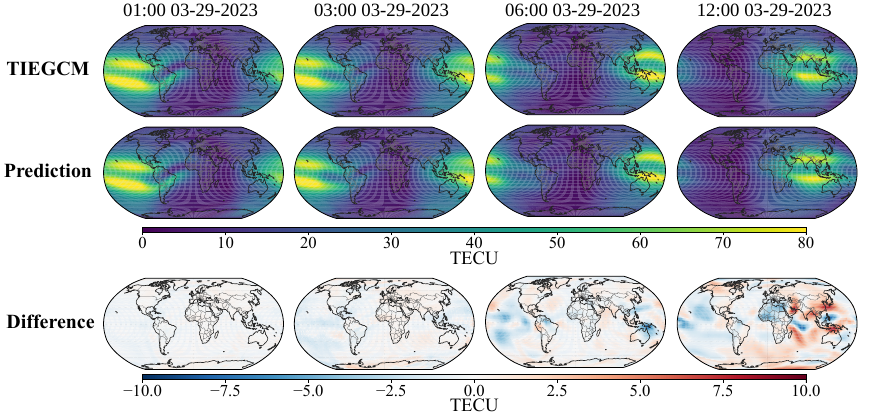}
\caption{Comparison of TIEGCM reference TEC maps and PGFM forecasts at lead times of 1, 3, 6, and 12~hours. The forecast was initialized at 00:00~UT on 29 March 2023.}
\label{fig:maps}
\end{figure}

Figure~\ref{fig:maps} shows the comparison between the reference TEC maps generated by TIEGCM and those predicted by the proposed model at lead times of 1, 3, 6, and 12~hours for the forecast initialized at 00:00~UT on 29 March 2023. In a broad view, the predicted maps reproduce the principal large-scale spatial features of the global TEC distribution, although the discrepancies become more noticeable as the forecast lead time increases.
At lead times of 1 and 3~hours, the differences are relatively small and spatially diffuse. The error increases at longer lead times, with the more pronounced differences becoming increasingly concentrated at low latitudes. This concentration is most evident at the 12-hour lead time, motivating a more detailed spatial analysis of the forecast errors.

\begin{figure}
\centering
\includegraphics[width=\textwidth]{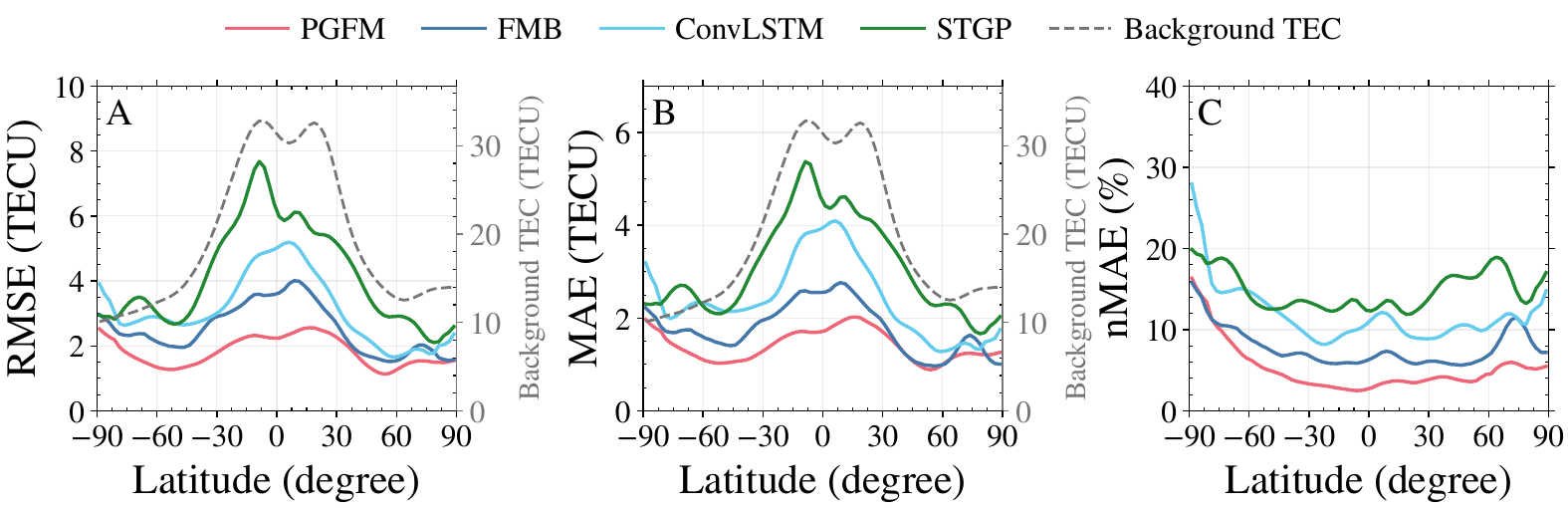}
\caption{Latitude-resolved profiles of forecast errors and background TEC. At each $2.5^\circ$ latitude, values were averaged over all longitudes. All profiles were smoothed using a five-point moving average for visualization. (A) RMSE, (B) MAE, and (C) nMAE. The gray dashed curves in (A) and (B) show background TEC on the right axes.}
\label{fig:latitude_metrics}
\end{figure}

The difference maps in Figure~\ref{fig:maps} indicate that the forecast errors have a clear latitudinal structure. Such spatial variability can be obscured by global averaging. Figure~\ref{fig:latitude_metrics} presents continuous latitude profiles of RMSE, MAE, and nMAE for the four deployable models with the corresponding background TEC.
The RMSE and MAE profiles in Figure~\ref{fig:latitude_metrics} (A) and (B) exhibit marked peaks at low latitudes, broadly coinciding with the region of enhanced background TEC. These peaks are particularly prominent for STGP and ConvLSTM and remain visible for FMB. In contrast, PGFM produces the lowest absolute errors across nearly the entire latitude range, with the largest reductions occurring at low latitudes, where the competing models exhibit their greatest errors.
The nMAE profiles exhibit a different latitudinal pattern. As shown in Figure~\ref{fig:latitude_metrics} (C), normalized errors are relatively low at low and middle latitudes but increase toward high latitudes. This contrast can be partly explained by the background TEC profiles shown by the gray dashed curves in Figure~\ref{fig:latitude_metrics} (A) and (B). Because nMAE normalizes the absolute error by the local TEC magnitude, the higher background TEC at low latitudes reduces the relative error despite the larger absolute errors in this region. In contrast, the lower background TEC at high latitudes amplifies the normalized error.
The increase in nMAE is larger toward the southern polar latitudes than toward the northern polar latitudes.
A possible physical explanation to this asymmetry is illustrated by Figures~\ref{fig:phys_weight}(C) and (D), where the southern polar region exhibits more distinct spatial variations in the net $\mathrm{O}^{+}$ column rate of change. Such spatial heterogeneity may make polar TEC evolution more difficult to predict.
PGFM maintains the lowest nMAE across most latitudes.

\subsection{Overall Forecasting Performance}
\label{subsec:overall_performance}

\begin{figure}[t]
\centering
\includegraphics[width=\textwidth]{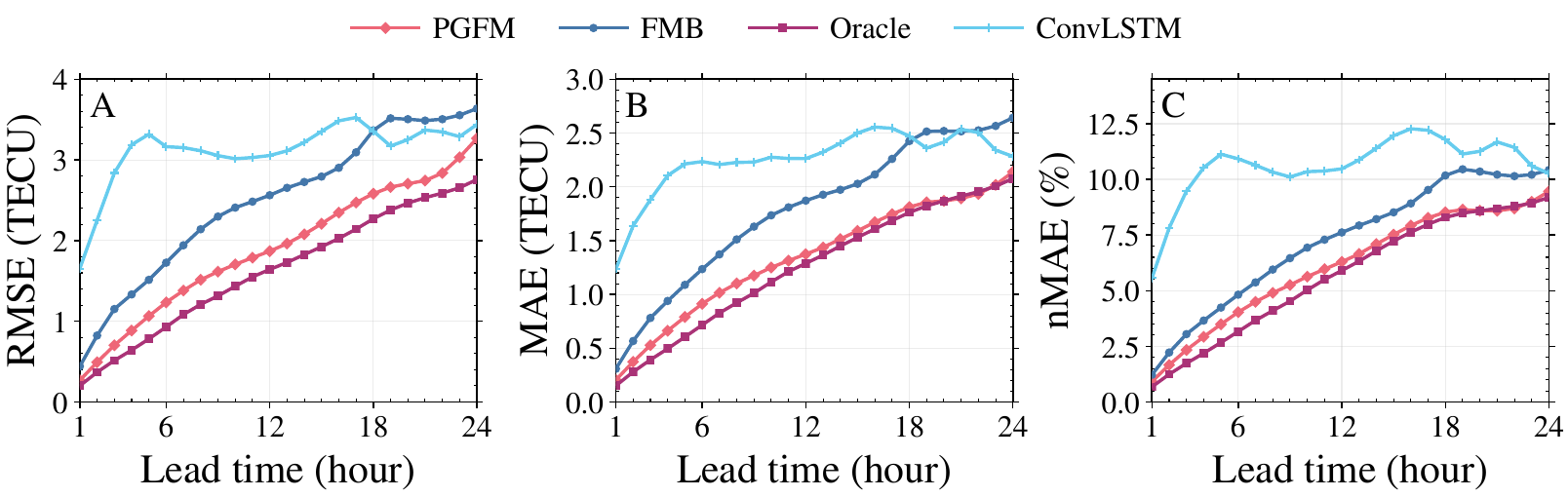}
\caption{Averaged forecasting errors of different models at lead times from 1 to 24~hours: (A) RMSE, (B) MAE, and (C) nMAE.}
\label{fig:leadtime_metrics}
\end{figure}

Table~\ref{tab:forecasting_by_horizon} reports RMSE, MAE, and nMAE at selected lead times for STGP, ConvLSTM, FMB, PGFM, and Oracle.
Figure~\ref{fig:leadtime_metrics} presents the error curves for ConvLSTM, FMB, PGFM, and Oracle throughout the 24-hour forecast horizon.
RMSE and MAE quantify absolute forecast errors in TECU, whereas nMAE expresses absolute errors relative to the overall TEC magnitude.
PGFM achieves the lowest errors among the deployable models for each metric and lead time shown.

\begin{table}[t]
\caption{Forecast errors at selected lead times for all evaluated models.}
\label{tab:forecasting_by_horizon}
\centering
\begin{tabular}{llccccc}
\hline
& & Statistical & Deep learning & \multicolumn{2}{c}{Flow matching} & Diagnostic \\
\hline
Metric & $h$ (hour) & STGP & ConvLSTM & FMB & PGFM & Oracle$^{a}$ \\
\hline
RMSE & 1 & 1.417 & 1.654 & \underline{0.442} & \textbf{0.269} & 0.201 \\
 & 3 & 3.078 & 2.838 & \underline{1.151} & \textbf{0.703} & 0.514 \\
 & 6 & 5.066 & 3.164 & \underline{1.726} & \textbf{1.234} & 0.929 \\
 & 12 & 4.829 & 3.054 & \underline{2.562} & \textbf{1.869} & 1.642 \\
 & 24 & 4.865 & \underline{3.439} & 3.636 & \textbf{3.266} & 2.757 \\
 & Avg. & 4.11 & 3.114 & \underline{2.482} & \textbf{1.892} & 1.623 \\
 & Imp. (\%) & 54.0 & 39.2 & 23.8 & -- & N/A \\
\hline
MAE & 1 & 1.034 & 1.235 & \underline{0.308} & \textbf{0.204} & 0.151 \\
 & 3 & 2.323 & 1.881 & \underline{0.781} & \textbf{0.527} & 0.392 \\
 & 6 & 3.558 & 2.234 & \underline{1.235} & \textbf{0.912} & 0.715 \\
 & 12 & 3.465 & 2.262 & \underline{1.871} & \textbf{1.374} & 1.288 \\
 & 24 & 3.617 & \underline{2.282} & 2.642 & \textbf{2.139} & 2.077 \\
 & Avg. & 3.425 & 2.250 & \underline{1.785} & \textbf{1.341} & 1.252 \\
 & Imp. (\%) & 60.9 & 40.4 & 24.9 & -- & N/A \\
\hline
nMAE & 1 & 4.594 & 5.561 & \underline{1.218} & \textbf{0.916} & 0.676 \\
 & 3 & 10.249 & 9.472 & \underline{3.047} & \textbf{2.335} & 1.736 \\
 & 6 & 15.762 & 10.923 & \underline{4.820} & \textbf{4.034} & 3.164 \\
 & 12 & 16.076 & 10.472 & \underline{7.616} & \textbf{6.295} & 5.903 \\
 & 24 & 16.009 & \underline{10.266} & 10.422 & \textbf{9.461} & 9.187 \\
 & Avg. & 13.038 & 10.607 & \underline{7.246} & \textbf{6.119} & 5.721 \\
 & Imp. (\%) & 53.1 & 42.3 & 15.5 & -- & N/A \\
\hline
\end{tabular}
\begin{tablenotes}
Values for ConvLSTM, FMB, PGFM, and Oracle were recovered from the plotted coordinates in Figure~\ref{fig:leadtime_metrics} and rounded to three decimal places.
STGP error values are retained from the earlier version of this table; its improvement percentages are recalculated using the updated PGFM averages.
RMSE and MAE are reported in TECU, and nMAE is reported in percent.
Avg. denotes the arithmetic mean over all 24 forecast lead times from 1 to 24~hours.
Imp. denotes the percentage reduction in average error achieved by PGFM relative to the corresponding comparator, calculated before rounding.
Bold values indicate the best deployable model.
Underlined values indicate the best deployable comparator other than PGFM.
$^{a}$Oracle uses true future physical information, which is unavailable in operational forecasting, and is excluded from model ranking and improvement calculations.
\end{tablenotes}
\end{table}

Averaged over all 24 lead times, PGFM achieves an RMSE of $1.892$~TECU, an MAE of $1.341$~TECU, and an nMAE of $6.119\%$.
Relative to ConvLSTM, these values represent reductions of $39.2\%$, $40.4\%$, and $42.3\%$, respectively.
Relative to the retained STGP results, the corresponding reductions are $54.0\%$, $60.9\%$, and $53.1\%$, respectively.
FMB also achieves lower average errors than ConvLSTM for all three metrics.
Within the flow matching framework, PGFM reduces average RMSE, MAE, and nMAE relative to FMB by $23.8\%$, $24.9\%$, and $15.5\%$, respectively.
PGFM maintains lower errors than FMB at every lead time, supporting the benefit of physical guidance in this comparison.

Figure~\ref{fig:leadtime_metrics} further shows how the relative performance of the models changes with forecast lead time.
Forecast errors generally increase as the lead time extends, reflecting the increasing difficulty of longer-range prediction.
From 1 to 12~hours, FMB is the strongest deployable comparator, while PGFM maintains lower errors for all three metrics.
ConvLSTM becomes more competitive at longer lead times and achieves lower errors than FMB for all three metrics at 24~hours.
At that lead time, PGFM achieves an RMSE of $3.266$~TECU, an MAE of $2.139$~TECU, and an nMAE of $9.461\%$.
The corresponding ConvLSTM errors are $3.439$~TECU, $2.282$~TECU, and $10.266\%$, respectively.
Although the performance margin varies with lead time, PGFM retains the lowest errors among the deployable models throughout the forecast horizon.

Oracle is included as a diagnostic reference rather than as a deployable forecasting model.
It uses the PGFM architecture but receives true future physical information during inference, allowing the comparison to assess the potential benefit of more accurate physical inputs.
Oracle achieves lower average errors than PGFM, with RMSEs of $1.623$ and $1.892$~TECU, MAEs of $1.252$ and $1.341$~TECU, and nMAEs of $5.721\%$ and $6.119\%$, respectively.
PGFM nevertheless improves on FMB without access to future physical information.
At longer lead times, the MAE and nMAE curves of PGFM and Oracle become close and occasionally cross, whereas Oracle maintains a lower RMSE throughout the horizon.
Because RMSE gives greater weight to large deviations, this pattern may indicate that accurate physical inputs help limit larger forecast errors.
Confirming this interpretation would require a targeted analysis of the error distribution.

\subsection{Analysis of the Learned ``Velocity"}
\label{subsec:velocity_physics}

\begin{figure}[t]
    \centering
    \includegraphics[width=\textwidth]{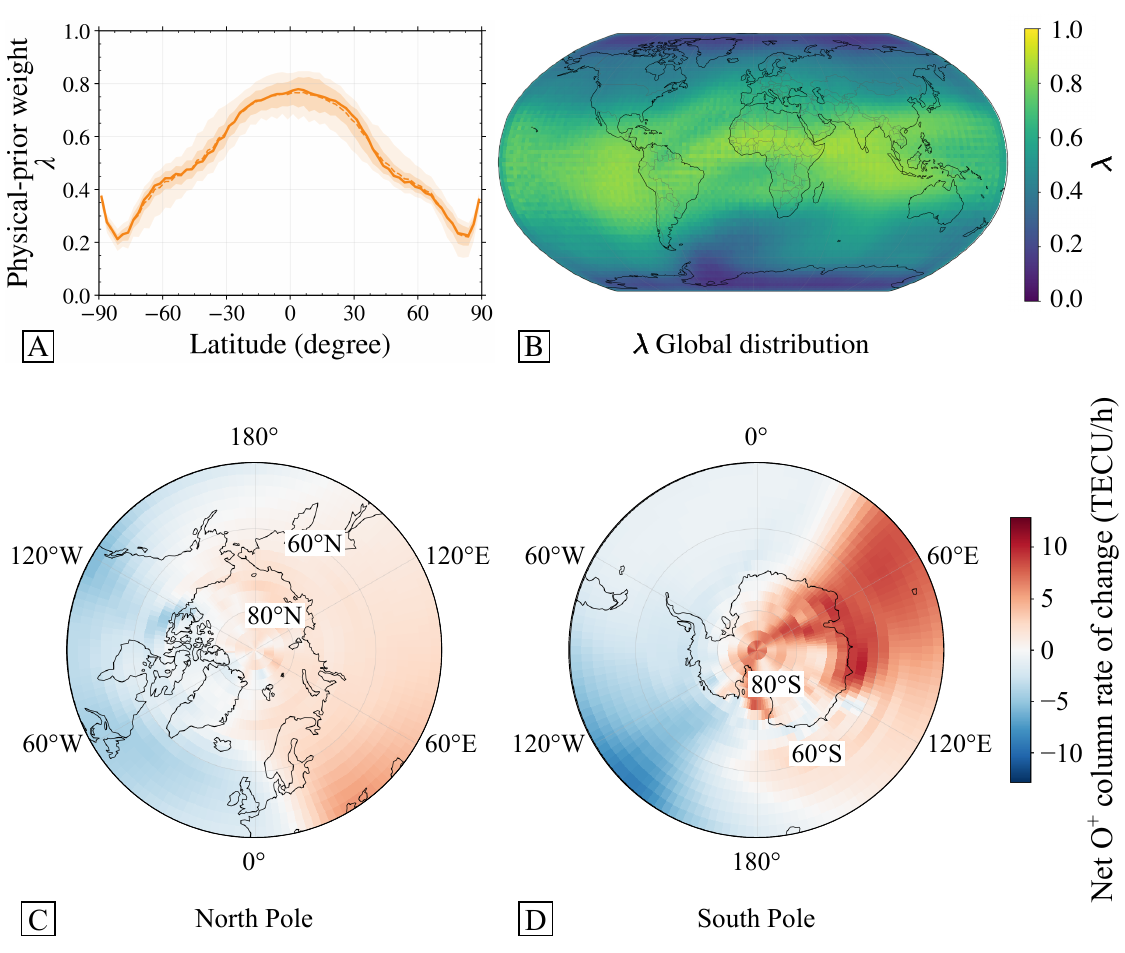}
    \caption{Learned fusion weight $\lambda$ assigned to the physical ``velocity". Larger values indicate a greater contribution of the physical branch to the fused ``velocity''. For each grid cell, $\lambda$ was first averaged over all test samples. (A) Latitudinal profiles of the resulting weights across longitude. The solid and dashed curves denote the median and mean, respectively, while the dark and light shaded regions indicate the interquartile range and the 10th--90th percentile interval. (B) Global distribution of the mean weight $\lambda$ at each grid cell. (C) and (D) Polar views of the height-integrated net $\mathrm{O}^{+}$ rate of change over the northern and southern polar regions, respectively, at 03:45~UTC on 3 September 2023. The rates are calculated by height-integrating the sum of production, loss, drift, and wind contributions and are expressed in $\mathrm{TECU \cdot h^{-1}}$ from TIEGCM outputs.}
    \label{fig:phys_weight}
\end{figure}

The previous spatial analysis shows that forecast difficulty varies with latitude. Figure~\ref{fig:phys_weight} examines how the proposed model balances the data-driven and physical ``velocity" through the fusion weight $\lambda$ defined in Equation~\eqref{eq:fused_velocity}. The weights are optimized jointly with the two estimators, with larger values assigning greater weight to the physical branch in predicting TEC changes. Figure~\ref{fig:phys_weight}(A) shows that $\lambda$ is generally larger at low and lower-middle latitudes and decreases toward middle and high latitudes. The region of larger weights overlaps the enhanced background TEC and the largest absolute reductions in RMSE and MAE in Figure~\ref{fig:latitude_metrics}.
Figure~\ref{fig:phys_weight} (B) shows the global distribution of the mean of $\lambda$.
This distribution is consistent with the latitudinal structure of the global average TEC distribution.
Low latitude TEC variations are more strongly influenced by relatively regular solar extreme ultraviolet forcing and organized electrodynamic processes, allowing the PDE-based tendency to provide a more predictable velocity contribution. In contrast, high-latitude TEC can respond rapidly to variable magnetospheric forcing and auroral precipitation \cite{lu2023understanding}, making subsequent physical tendencies more difficult to predict. The smaller \(\lambda\) values at higher latitudes therefore suggest that the model relies more heavily on the data-driven estimator where the physics-based tendency is less reliable at the available resolution. At the same time, the spatial pattern of \(\lambda\) broadly follows the magnetic-equator region, which is consistent with the influence of \(E\times B\) drift and the equatorial ionization anomaly on TEC variability.
Figures~\ref{fig:phys_weight}(C) and (D) provide physical context by showing polar distributions of the height-integrated net $\mathrm{O}^{+}$ rate of change. The southern polar region exhibits the noticeable dynamics and complex spatial variations in these rates. TIEGCM outputs sampled at 15-min intervals may not fully capture the temporal variability of physical process rates in polar regions, potentially limiting the information available for training the physical estimator.
These results demonstrate that the model learns spatially varying weights for combining physical with data-driven ``velocity'' estimates, which a single global fusion coefficient would not represent.

\begin{figure}[t]
    \centering
    \includegraphics[width=\textwidth]{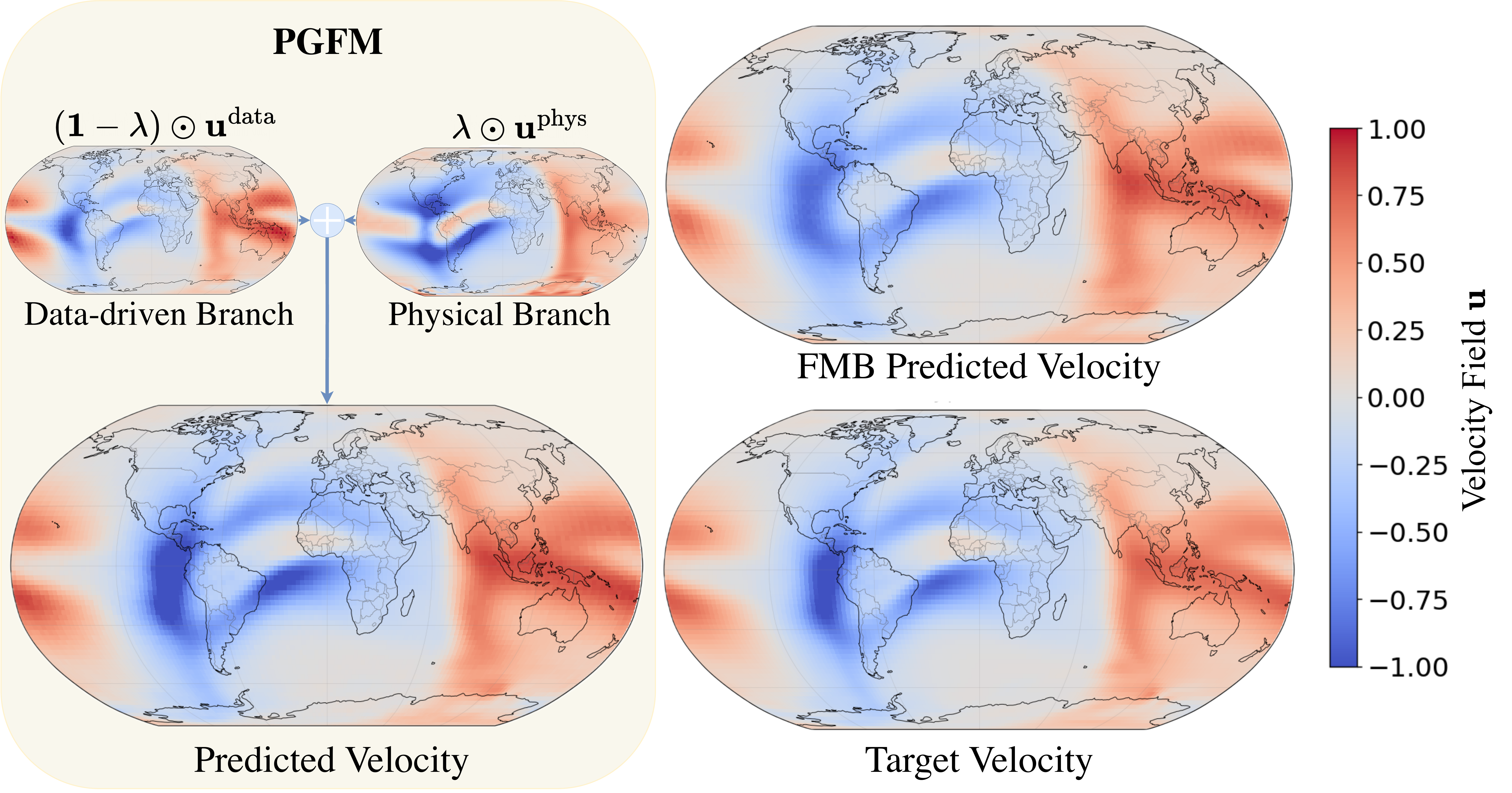}
    \caption{Predicted flow matching ``velocity" maps at 03:00~UT on 29 August 2023.
    The panels compare the target ``velocity", the outputs of the data-driven and physical estimators, the fused PGFM prediction, and the FMB prediction. The ``velocity" describes changes in the flow matching state space; positive and negative values indicate tendencies toward increasing and decreasing TEC values, respectively.
    }
    \label{fig:velocity_branch_maps}
\end{figure}

Figure~\ref{fig:velocity_branch_maps} provides a view of the adaptive fusion by comparing the target ``velocity" with our prediction containing the data-driven and physical estimates, and the prediction of the FMB.
In flow matching, each global TEC map is treated as a high-dimensional state, and the associated ``velocity" has the same dimensions as the TEC map, with one scalar component for each spatial grid point.  Figure~\ref{fig:velocity_branch_maps} shows these components at their corresponding locations on the latitude--longitude grid. Along the flow matching path, a positive component moves the TEC value at that location toward larger values, whereas a negative component moves it toward smaller values. The displayed quantity is a ``velocity" in the space of flow matching, rather than a physical plasma drift velocity.
The figure shows the predicted ``velocity" snapshot at 03:00~UT on 29 August 2023. The target field exhibits organized low-latitude structures, with longitudinal differences that may partly reflect the contrast between sunlit and nighttime sectors.
In the proposed model, the data-driven and physical branches capture different aspects of the target spatial structure, indicating that the two sources provide complementary dynamical information. After adaptive fusion, the predicted ``velocity" closely reproduces the general spatial pattern of the target field. More importantly, compared with the FMB, our model better recovers the spatial extent of the structures in the target ``velocity".
This comparison shows that the adaptive integration of physical ``dynamics" helps the learned transport dynamics toward a ``velocity" that is more consistent with the physical evolution.

Overall, PGFM outperforms all selected benchmarks by combining a novel generative framework with a physics-guided design. During training, the probability path we designed provides informative physical guidance for learning physically meaningful transport dynamics. During forecasting, the model does not assign a fixed contribution to the physical ``velocity". It dynamically balances the physical and data-driven branches through spatially adaptive weighting. This design yields consistently accurate and physically interpretable ionospheric forecasts across varying forecast regions.

\section{Conclusion}
\label{sec:conclusion}

Accurate global TEC forecasting is important for characterizing ionospheric variability and supporting space weather applications. We have proposed a physics-guided flow matching framework that constructs a probability path directly following the evolution prescribed by an ionospheric PDE as ``physical flow". In addition, the \(\mathrm O^+\) continuity equation provides a physical ``velocity'' estimate that is combined with a data-driven ``velocity'' estimate through a spatially adaptive fusion mechanism.
Evaluated using the dataset generated by TIEGCM, our model demonstrates substantial improvements in both spatial and temporal forecasting accuracy compared with representative forecasting benchmarks.
The learned weighting fields show that the physical contribution varies spatially. Analysis of the branch velocities further shows how adaptive fusion changes the learned transport field and aligns with the target structure.
Future research will investigate computationally efficient latent space formulations designed to reduce computational and memory costs while preserving the physical guidance.
The proposed framework may also be extended to other spatiotemporal systems whose evolution is governed by known PDEs. Bridging generative flow matching with physical domain knowledge opens up a new avenue for physically consistent forecasting in complex geoscientific systems.

\section*{Conflict of Interest declaration}
The authors declare there are no conflicts of interest for this manuscript.

\acknowledgments
This work has been supported in part by the U.S. National Science Foundation (NSF) under awards AGS-2149695, AGS-1753214, CEDAR-2431688, and AGS-2347149, and by the U.S. National Aeronautics and Space Administration (NASA) under awards 80NSSC22K0018, 80NSSC22K1010, 80NSSC26K0295, and 80NSSC22M0064.

\bibliography{reference}

\appendix

\section{Flow Matching Background}
\label{app:fm_background}

Flow matching learns a time-dependent vector field $v_{\theta}(\mathrm z,\tau)$ whose associated flow transports samples from a prior distribution $\pi_0$ to a target distribution $\pi_1$ \cite{lipman2023flow}. Let $\mathcal{Z}\subseteq\mathbb{R}^{d}$ denote the state space and $\tau\in[0,1]$ a normalized flow coordinate. A single state can represent a complete gridded physical field. The state evolves according to
\begin{equation}
    \frac{\mathrm{d}\mathrm z_{\tau}}{\mathrm{d}\tau}
    =
    v_{\theta}(\mathrm z_{\tau},\tau),
    \qquad
    \mathrm z_0\sim\pi_0.
    \label{eq:app_fm_ode}
\end{equation}
The solution of this ODE defines a flow map $\phi_{\theta,\tau}$ such that $\mathrm z_{\tau}=\phi_{\theta,\tau}(\mathrm z_0)$. The distribution at $\tau$ is the pushforward $p_{\theta,\tau}=(\phi_{\theta,\tau})_{\#}\pi_0$, where $(\cdot)_{\#}$ denotes the pushforward operator. The objective is to learn a vector field for which the terminal distribution satisfies $p_{\theta,1}\approx\pi_1$.

Training is formulated through a prescribed probability density path $\{p_{\tau}\}_{\tau\in[0,1]}$ connecting $p_0=\pi_0$ and $p_1=\pi_1$. Its associated target ``velocity field" $u_{\tau}(\mathrm z)$ satisfies the continuity equation
\begin{equation}
    \frac{\partial p_{\tau}(\mathrm z)}{\partial \tau}
    +
    \nabla_{\mathrm z}\cdot
    \left(
        p_{\tau}(\mathrm z)u_{\tau}(\mathrm z)
    \right)
    =
    0.
    \label{eq:app_fm_continuity}
\end{equation}
This equation ensures that the probability mass transported by $u_{\tau}$ follows the prescribed density path. In practice, however, the marginal density $p_{\tau}$ and its corresponding ``velocity field" are generally not available in closed form.

Flow matching addresses this difficulty by constructing tractable conditional paths between endpoint pairs. Let $(\mathrm z_0,\mathrm z_1)\sim\gamma$, where $\gamma$ is a coupling with marginals $\pi_0$ and $\pi_1$. A general Gaussian conditional path can be written as
\begin{equation}
    p_{\tau}(\mathrm z\mid \mathrm z_0,\mathrm z_1)
    =
    \mathcal{N}
    \left(
        \mathrm z
        \,\middle|\,
        a_{\tau}\mathrm z_0+b_{\tau}\mathrm z_1,\,
        c_{\tau}^{2}\mathbf{I}
    \right),
    \qquad
    \mathrm z_{\tau}
    =
    a_{\tau}\mathrm z_0+b_{\tau}\mathrm z_1+c_{\tau}\epsilon,
    \quad
    \epsilon\sim\mathcal{N}(0,\mathbf{I}).
    \label{eq:app_conditional_path}
\end{equation}
Here, $a_{\tau}$, $b_{\tau}$, and $c_{\tau}$ control the evolution of the conditional mean and variance. They are assumed to be continuous on $[0,1]$ and differentiable on $(0,1)$, with $c_{\tau}>0$ in the interior of the flow interval. Their endpoint values specify the prescribed source and target distributions. The marginal probability path is obtained by averaging $p_{\tau}(\mathrm z\mid \mathrm z_0,\mathrm z_1)$ over endpoint pairs drawn from $\gamma$.

Holding $(\mathrm z_0,\mathrm z_1,\epsilon)$ fixed, differentiation of Equation~\eqref{eq:app_conditional_path} gives the conditional target ``velocity''
\begin{equation}
    u_{\tau}(\mathrm z_{\tau}\mid \mathrm z_0,\mathrm z_1)
    =
    \frac{\mathrm{d}\mathrm z_{\tau}}{\mathrm{d}\tau}
    =
    \dot{a}_{\tau}\mathrm z_0
    +
    \dot{b}_{\tau}\mathrm z_1
    +
    \dot{c}_{\tau}\epsilon,
    \label{eq:app_conditional_velocity}
\end{equation}
where the dot denotes differentiation with respect to $\tau$. The neural vector field is then trained using the conditional flow-matching objective
\begin{equation}
    \mathcal{L}_{\mathrm{FM}}(\theta)
    =
    \mathbb{E}_{
        \substack{
            \tau\sim\mathcal{U}[0,1],\,
            (\mathrm z_0,\mathrm z_1)\sim\gamma,\\
            \epsilon\sim\mathcal{N}(0,\mathbf{I})
        }
    }
    \left[
        \left\|
            v_{\theta}(\mathrm z_{\tau},\tau)
            -
            u_{\tau}(\mathrm z_{\tau}\mid \mathrm z_0,\mathrm z_1)
        \right\|_2^2
    \right].
    \label{eq:app_fm_loss}
\end{equation}
This conditional regression avoids explicit evaluation of the marginal density and its ``velocity field" while providing the flow-matching training gradient \cite{lipman2023flow}. After training, samples from $\pi_0$ are transported toward $\pi_1$ by integrating Equation~\eqref{eq:app_fm_ode} from $\tau=0$ to $\tau=1$. The particular coupling and path parameters used in the proposed model are specified in the main text.

\section{Network Architecture}
\label{app:networks}

This section describes the overall architecture for ``velocity" estimation, which consists of two components: a data-driven``velocity''estimator and a physical``velocity''estimator. The following subsections describe these components and their combination into the final predicted ``velocity".

\begin{figure}[htbp]
    \centering
    \includegraphics[width=\linewidth]{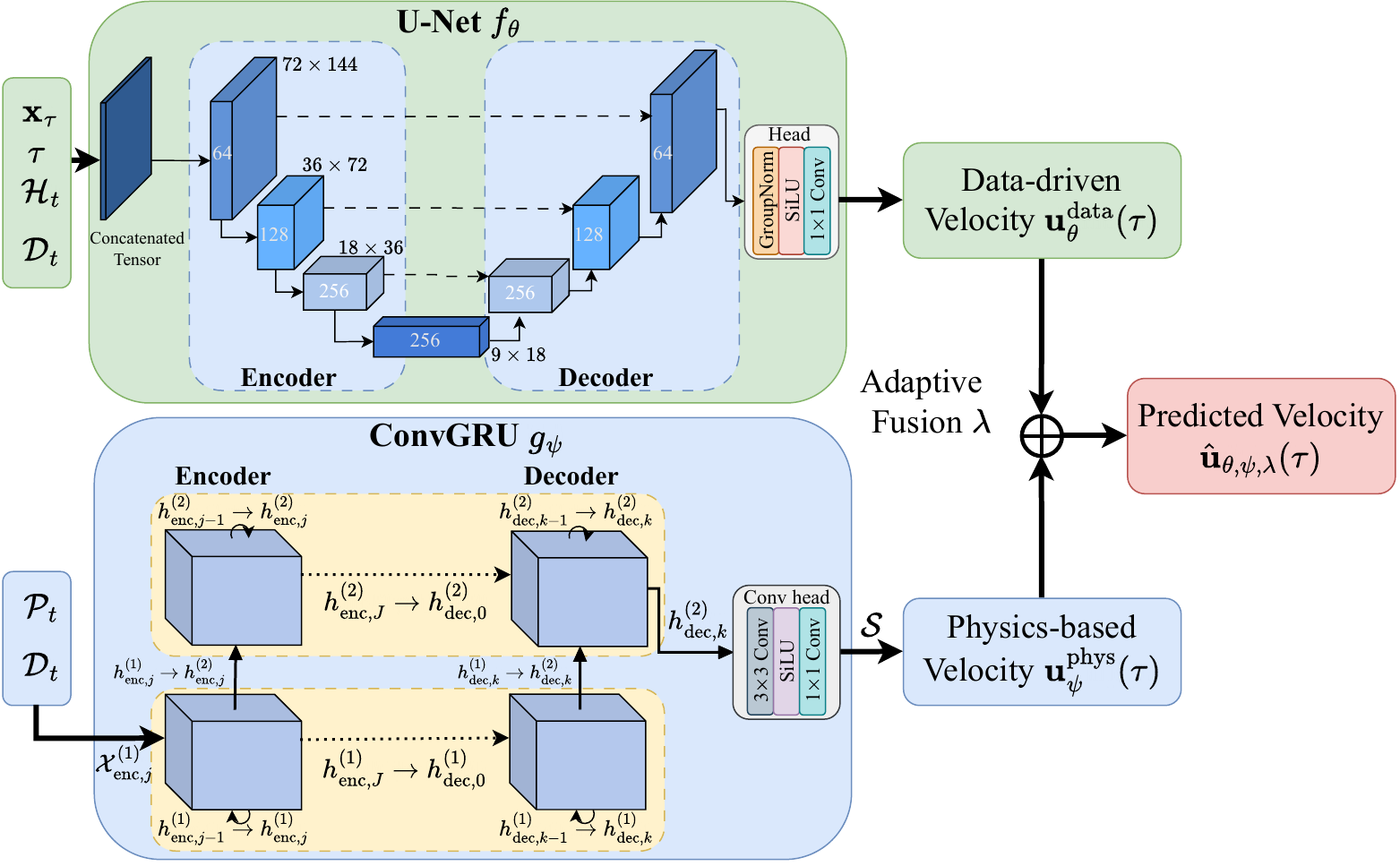}
    \caption{
    Architecture of the data-driven and physical``velocity''estimators. The conditional U-Net takes the intermediate TEC state $\mathrm x_\tau$, the flow coordinate $\tau$,  encoded past TEC history $\mathcal{H}_t$ and encoded physical-driver history $\mathcal{D}_t$ as inputs to estimate the data-driven ``velocity''. The ConvGRU predicts a physical basis of the ``velocity" by summarizing the past ion dynamics $\mathcal P_t$ and available physical drivers $\mathcal D_t$.
    Then the two estimates are combined at each grid cell adaptively to generate the predicted ``velocity".}
    \label{fig:networks}
\end{figure}

\subsection{Data-Driven``velocity''Estimator}
\label{app:data_velocity_network}

The data-driven estimator uses a two-dimensional U-Net \cite{ronneberger2015unet} with residual blocks \cite{he2015deep}. It follows an encoder-decoder architecture with skip connections between corresponding spatial resolutions. The residual blocks incorporate group normalization \cite{wu2018group} and sigmoid-weighted linear unit (SiLU) activations \cite{elfwing2018sigmoid}.

The network uses standardized inputs. Let \(m_{\mathrm{TEC}}\) and \(\sigma_{\mathrm{TEC}}>0\) denote the mean and standard deviation of TEC estimated from the training data, respectively. The intermediate flow state
\(\mathbf{x}\) is standardized as
\(
    {\mathbf{x}}
    =
    \frac{\mathbf{X}-m_{\mathrm{TEC}}}{\sigma_{\mathrm{TEC}}},
\)
where the subtraction and division are applied elementwise. For forecasting in these standardized coordinates, the initial state is $\mathbf{x}_0=(\mathbf{X}_t-m_{\mathrm{TEC}})/\sigma_{\mathrm{TEC}}$, and the terminal state is transformed back to TEC units as $\widehat{\mathbf{X}}_{t+1}=m_{\mathrm{TEC}}+\sigma_{\mathrm{TEC}}\mathbf{x}_1$. Physical covariates are standardized separately using their training-data statistics. Small positive constants are added to standard deviations in the implementation to avoid division by zero.

The physical covariates comprise eight scalar channels and three spatial channels. The scalar channels contain the planetary geomagnetic index, daily and averaged solar radio flux, hemispheric power, cross-polar-cap potential, and three interplanetary magnetic-field components. The spatial channels contain the lower-boundary temperature and two horizontal wind components.
The historical TEC sequence \(\mathcal H_t\) is standardized and processed by two \(3\times3\times3\) convolutions, each followed by a SiLU activation. The convolutions produce 16 channels and preserve the temporal and spatial dimensions. Averaging over the temporal dimension gives \(\mathbf A_t\in\mathbb R^{16\times H\times W}\), which represents the encoded TEC history.
The physical-driver history \(\mathcal D_t\) contains the scalar and spatial covariates described above. Its three spatial channels are processed by a separate pair of \(3\times3\times3\) convolutions with SiLU activations. Temporal averaging gives \(\mathbf B_t\in\mathbb R^{12\times H\times W}\), which represents the encoded history of the lower-boundary temperature and two horizontal wind components.
The eight scalar channels of \(\mathcal D_t\) are processed by two one-dimensional convolutions with a kernel size of 3 and SiLU activations. Temporal averaging produces an eight-dimensional feature vector, which is repeated at every spatial grid cell to obtain \(\mathbf C_t\in\mathbb R^{8\times H\times W}\). Thus, \(\mathbf C_t\) represents the encoded scalar-driver history and has the same feature values at every grid cell.
All context convolutions use a stride of 1 and padding of 1. The subscript \(t\) indicates the forecast initialization time at which the historical inputs end.

Let \(\mathbf 1\in\mathbb R^{H\times W}\) denote the all-ones map, and let \(\operatorname{Concat}\) denote concatenation along the channel dimension. The U-Net input is
\begin{equation}
    \mathbf Q_{\tau}
    =
    \operatorname{Concat}
    \left(
        {\mathbf x}_{\tau},
        \tau\mathbf 1,
        \mathbf A_t,
        \mathbf B_t,
        \mathbf C_t
    \right)
    \in\mathbb R^{38\times H\times W},
    \label{eq:app_unet_input}
\end{equation}
The intermediate state and the flow-coordinate map each contribute one channel, while the historical features contribute 16, 12, and 8 channels, respectively.

An initial \(3\times3\) convolution maps the 38-channel input to 64 channels. The encoder consists of three stages with 64, 128, and 256 output channels, respectively. Each stage contains two residual blocks, followed by a \(4\times4\) convolution with stride 2 and padding 1 that halves the spatial resolution. The bottleneck then processes the features using two residual blocks with 256 channels. The decoder progressively restores the spatial resolution through three stages. Each stage begins with a \(4\times4\) transposed convolution with stride 2 and padding 1 that doubles the spatial resolution. The upsampled features are concatenated with the encoder features at the same resolution and passed through two residual blocks. The three decoder stages produce 256, 128, and 64 output channels, respectively.
The architecture is illustrated in the upper-left block of Figure~\ref{fig:networks}.

For a feature tensor \(\mathbf a\), a residual block has the form shown in Equation~\eqref{eq:app_residual_block}:
\begin{equation}
    \operatorname{RB}(\mathbf a)
    =
    \mathcal C_2
    \left(
        \operatorname{SiLU}
        \left[
            \operatorname{GN}_2
            \left(
                \mathcal C_1
                \left[
                    \operatorname{SiLU}
                    \left(
                        \operatorname{GN}_1(\mathbf a)
                    \right)
                \right]
            \right)
        \right]
    \right)
    +
    \mathcal P(\mathbf a),
    \label{eq:app_residual_block}
\end{equation}
where \(\mathcal C_1\) and \(\mathcal C_2\) are \(3\times3\) convolutions with stride 1 and padding 1, and \(\operatorname{GN}_1\) and \(\operatorname{GN}_2\) are group-normalization layers with eight groups. The shortcut \(\mathcal P\) is the identity when the input and output channel counts agree and a \(1\times1\) convolution otherwise. The configuration described here uses no dropout. The activation is \(\operatorname{SiLU}(\mathbf a)=\mathbf a/(1+\exp(\mathbf{-a}))\), applied elementwise.

The final decoder output has 64 channels and is processed by group normalization with eight groups, followed by SiLU. A \(1\times1\) convolution then maps these features to a single-channel standardized ``velocity" \({\mathbf u}^{\mathrm{data}}_{\theta} \in\mathbb R^{H\times W}\).

\subsection{Physical``velocity''Estimator}
\label{app:physics_velocity_network}

The physical estimator is implemented as a Convolutional Gated Recurrent Unit (ConvGRU) \cite{ballas2016delving}. It predicts the net TEC tendency over one forecast cycle from the historical physical-process rates. The predicted dynamics are then mapped to the flow space to obtain $\mathbf u^{\mathrm{phys}}_{\psi}(\tau)$. The architecture is illustrated in the lower-left block of Figure~\ref{fig:networks}.

The estimator is conditioned on the historical ionospheric dynamics $\mathcal P_t$ and physical-driver history $\mathcal D_t$. The sequence $\mathcal P_t$ contains four dynamical terms associated with $\mathrm{O}^{+}$ production, chemical loss, electrodynamic drift, and neutral-wind transport. These terms characterize the physical processes governing ion-density evolution and provide a physical basis for estimating TEC dynamics. The sequence $\mathcal D_t$ contains the eight scalar and three spatial covariates described in Section~\ref{app:data_velocity_network}. Both historical sequences end at the forecast initialization time $t$. Future physical dynamics are predicted by the network and are not required as conditioning inputs.

The encoder contains two stacked ConvGRU layers. Let $J$ denote the number of historical input steps supplied to the ConvGRU encoder, and let $\boldsymbol{\chi}_{\mathrm{enc},j}^{(1)}$ denote the physical input assembled from the corresponding elements of $\mathcal P_t$ and $\mathcal D_t$ at historical step $j$. The encoder processes these inputs in chronological order. Its hidden states are updated as
\begin{equation}
    \begin{aligned}
        \mathbf h_{\mathrm{enc},j}^{(1)}
        &=
        \operatorname{ConvGRU}_{\mathrm{enc}}^{(1)}
        \left(
            \boldsymbol{\chi}_{\mathrm{enc},j}^{(1)},
            \mathbf h_{\mathrm{enc},j-1}^{(1)}
        \right), \\
        \mathbf h_{\mathrm{enc},j}^{(2)}
        &=
        \operatorname{ConvGRU}_{\mathrm{enc}}^{(2)}
        \left(
            \mathbf h_{\mathrm{enc},j}^{(1)},
            \mathbf h_{\mathrm{enc},j-1}^{(2)}
        \right),
        \qquad j=1,\ldots,J.
    \end{aligned}
    \label{eq:app_physics_encoder}
\end{equation}
Each ConvGRU layer combines the current input with its previous hidden state through convolutional gating operations. The first layer processes the physical input sequence, while the second layer processes the evolving features produced by the first layer. The final hidden states summarize the spatial and temporal information in the available physical history.

The decoder also contains two stacked ConvGRU layers. Each decoder layer is initialized with the final hidden state of the corresponding encoder layer:
\(
    \mathbf h_{\mathrm{dec},0}^{(\ell)}
    =
    \mathbf h_{\mathrm{enc},J}^{(\ell)},
    \, \ell=1,2 .
\)
Starting from these hidden states, the decoder
recurrently updates its hidden representations to produce features
for the future physical dynamics. Let $K$ denote the number of
prediction steps within one forecast cycle. The decoder updates
are written as
\begin{equation}
    \begin{aligned}
        \mathbf h_{\mathrm{dec},k}^{(1)}
        &=
        \operatorname{ConvGRU}_{\mathrm{dec}}^{(1)}
        \left(
            \mathbf h_{\mathrm{dec},k-1}^{(1)}
        \right), \\
        \mathbf h_{\mathrm{dec},k}^{(2)}
        &=
        \operatorname{ConvGRU}_{\mathrm{dec}}^{(2)}
        \left(
            \mathbf h_{\mathrm{dec},k}^{(1)},
            \mathbf h_{\mathrm{dec},k-1}^{(2)}
        \right),
        \qquad k=1,\ldots,K.
    \end{aligned}
    \label{eq:app_physics_decoder}
\end{equation}
The first layer evolves its previous hidden state, while the second layer combines the current first-layer features with its own previous hidden state.

At each decoder step \(k\), the second-layer hidden state \(\mathbf h_{\mathrm{dec},k}^{(2)}\) is passed through the convolutional output head:
\begin{equation}
\widehat{\mathbf r}_{\psi,k}
=
\mathcal C_{\mathrm{head},1}
\left(
    \operatorname{SiLU}
    \left[
        \mathcal C_{\mathrm{head},3}
        \left(
            \mathbf h_{\mathrm{dec},k}^{(2)}
        \right)
    \right]
\right),
\qquad k=1,\ldots,K,
\label{eq:app_physics_output_head}
\end{equation}
where \(\mathcal C_{\mathrm{head},3}\) is a \(3\times3\) convolution with 32 output channels and \(\mathcal C_{\mathrm{head},1}\) is a \(1\times1\) convolution with one output channel. The output \(\widehat{\mathbf r}_{\psi,k}\in\mathbb R^{H\times W}\) is a single map of the predicted net TEC tendency at decoder step \(k\), representing the combined contribution of the physical processes. It is expressed in physical units of TECU per second and is distinct from the separate historical process channels in $\mathcal P_t$.
The ConvGRU forecaster produces the sequence
\(
\left\{
    \widehat{\mathbf r}_{\psi,k}
\right\}_{k=1}^{K}
=
g_\psi
\left(
    \mathcal P_t,
    \mathcal D_t
\right)
\),
where \(K\) is the number of predictions produced over the forecast interval.

\subsection{Adaptive``velocity''Fusion}
\label{app:adaptive_velocity_fusion}

To combine the two branches, their outputs must first be expressed at the same flow coordinate and in the same``velocity''space. The physical estimator produces a sequence of $K$ physical predictions over one forecast cycle. We associate these predictions with $K$ uniform subintervals of the flow interval $[0,1]$. For a flow coordinate $\tau\in[0,1]$, the corresponding prediction
is selected using
\begin{equation}
    \kappa(\tau)
    =
    \min\left\{
        \left\lfloor K\tau\right\rfloor+1,
        K
    \right\},
    \label{eq:app_flow_physics_index}
\end{equation}
where $\lfloor\cdot\rfloor$ denotes the floor function and $\kappa(1)=K$. The selected physical prediction is then mapped to the normalized flow space:
\begin{equation}
    \mathbf u_{\psi}^{\mathrm{phys}}(\tau)
    =
    \mathcal S
    \left(
        \widehat{\mathbf r}_{\psi,\kappa(\tau)}
    \right)
    =\frac{\Delta t}{\sigma_{\mathrm{TEC}}}
    \widehat{\mathbf r}_{\psi,\kappa(\tau)},
    \label{eq:app_aligned_physics_velocity}
\end{equation}
where $\Delta t$ is the physical forecast interval, expressed in the time unit used for the predicted tendency, and $\sigma_{\mathrm{TEC}}$ is the training-data TEC standard deviation defined in Section~\ref{app:data_velocity_network}. The factor follows from $\mathbf x=(\mathbf X-m_{\mathrm{TEC}})/\sigma_{\mathrm{TEC}}$ and the time relation $t+\tau\Delta t$. For a one-hour transition and tendencies expressed in TECU per second, $\Delta t=3600$ seconds. This conversion places the physical and data-driven estimates in the same``velocity''space at the same flow coordinate.
The two``velocity''estimates are combined through a learnable spatial weighting map
$\boldsymbol{\lambda}(\tau)\in[0,1]^{H\times W}$, whose entry $\lambda_{ij}(\tau)$ is the weight at grid cell $(i,j)$ and flow coordinate $\tau$:
\begin{equation}
    \widehat{\mathbf u}_{\theta,\psi,\lambda}(\tau)
    =
    \left(
        \mathbf 1-\boldsymbol{\lambda}(\tau)
    \right)
    \odot
    \mathbf u_{\theta}^{\mathrm{data}}(\tau)
    +
    \boldsymbol{\lambda}(\tau)
    \odot
    \mathbf u_{\psi}^{\mathrm{phys}}(\tau),
    \label{eq:app_adaptive_velocity_fusion}
\end{equation}
where $\mathbf 1\in\mathbb R^{H\times W}$ is the all-ones map and $\odot$ denotes elementwise multiplication. At each spatial location, $\boldsymbol{\lambda}(\tau)$ controls the contribution of the physical estimate, while $\mathbf 1-\boldsymbol{\lambda}(\tau)$ controls that of the data-driven estimate.

During training, the objective of flow matching in Equation~\eqref{eq:pgfm_training_objective} is applied to the fused ``velocity''. The fusion weights are learned to reduce the discrepancy between this final estimate and the target ``velocity''.
\end{document}